%% file: paper.tex
\documentclass[sigconf]{acmart}

\acmSubmissionID{1078}

\input{packages}
\input{copyright}
\input{definitions}
\input{authors}

\begin{document}
\title{Cascading Hybrid Bandits: Online Learning to Rank for Relevance and Diversity}

\begin{abstract} 
Relevance ranking and result diversification are two core areas in modern recommender systems.  
Relevance ranking aims at building a ranked list sorted in decreasing order of item relevance, while result diversification focuses on generating a ranked list of items that covers a broad range of topics. 
In this paper, we study an online learning setting that aims to recommend a ranked list with $K$ items that maximizes the ranking utility, i.e., a list whose items are relevant and whose topics are diverse. 
We formulate it as the \emph{cascade hybrid bandits}~(\acs{CHB})\acused{CHB} problem.  
\ac{CHB} assumes the cascading user behavior, where a user browses the displayed list  from top to bottom, clicks the first attractive item, and stops browsing the rest. 
We propose a hybrid contextual bandit approach, called $\cascadehybrid$, for solving this problem.
$\cascadehybrid$ models item relevance and topical diversity using two independent functions and simultaneously learns those functions from user click feedback.
We conduct experiments to evaluate $\cascadehybrid$ on two real-world recommendation datasets: MovieLens and Yahoo music datasets.
Our experimental results show that $\cascadehybrid$ outperforms the baselines.
In addition, we prove theoretical guarantees on the $n$-step performance demonstrating the soundness of $\cascadehybrid$.
\end{abstract}
 
%
%



\maketitle

\acresetall

\input{sections/1-Introduction}
\input{sections/2-Background}
\input{sections/3-Algorithm}
\input{sections/4-Experiments}

\input{sections/5-Analysis}
\input{sections/6-RelatedWork}
\input{sections/7-Conclusions}

\begin{acks}
This research was partially supported by the Netherlands Organisation for Scientific Research (NWO) under pro\-ject nr
612.\-001.\-551. 
All content represents the opinion of the authors, which is not necessarily shared or endorsed by their respective employers and/or sponsors.
\end{acks}

\bibliographystyle{ACM-Reference-Format}
\bibliography{Bibliography}

\end{document}

%% file: packages.tex
\usepackage{acronym}

\usepackage{tabu}
\usepackage{color}
\usepackage{algpseudocode}
\usepackage{algorithm}
\usepackage{amsbsy}
\usepackage{amsmath}
\usepackage{bm}
\usepackage{dsfont}
\usepackage[all]{xy}
\usepackage[capitalize]{cleveref}
\usepackage[inline]{enumitem}

\usepackage{graphicx}
\usepackage{epstopdf}

%% file: copyright.tex
\copyrightyear{2020}
\acmYear{2020}
\setcopyright{acmlicensed}
\acmConference[RecSys '20]{Fourteenth ACM Conference on Recommender Systems}{September 22--26, 2020}{Virtual Event, Brazil}
\acmBooktitle{Fourteenth ACM Conference on Recommender Systems (RecSys '20), September 22--26, 2020, Virtual Event, Brazil}
\acmPrice{15.00}
\acmDOI{10.1145/3383313.3412245}
\acmISBN{978-1-4503-7583-2/20/09}

%% file: definitions.tex
\usepackage{color}
\definecolor{Blue}{rgb}{0.9,0.3,0.3}
\usepackage{cancel}

\newcommand{\real}{\mathbb{R}}

\newcommand{\myvec}[1]{\mathbf{#1}}
\newcommand{\myvecsym}[1]{\boldsymbol{#1}}
\newcommand{\ind}[1]{\mathds{1}(#1)}

\newcommand{\valpha}{\myvecsym{\alpha}}
\newcommand{\vbeta}{\myvecsym{\beta}}

\newcommand{\vomega}{\myvecsym{\omega}}

\newcommand{\vphi}{\myvecsym{\phi}}

\newcommand{\vtheta}{\myvecsym{\theta}}

\newcommand{\vu}{\myvec{u}}
\newcommand{\vw}{\myvec{w}}

\newcommand{\vx}{\myvec{x}}

\newcommand{\vy}{\myvec{y}}

\newcommand{\vz}{\myvec{z}}

\newcommand{\vA}{\myvec{A}}
\newcommand{\vB}{\myvec{B}}

\newcommand{\vF}{\myvec{F}}
\newcommand{\vG}{\myvec{G}}
\newcommand{\vH}{\myvec{H}}
\newcommand{\vI}{\myvec{I}}

\newcommand{\vM}{\myvec{M}}

\newcommand{\vO}{\myvec{O}}

\newcommand{\expect}[1]{\mathbb{E}\left[ {#1} \right]}

\newcommand{\be}{\begin{equation}}
\newcommand{\ee}{\end{equation}}
\newcommand{\bea}{\begin{eqnarray}}
\newcommand{\eea}{\end{eqnarray}}
\newcommand{\beaa}{\begin{eqnarray*}}
\newcommand{\eeaa}{\end{eqnarray*}}

\DeclareMathAlphabet{\mathpzc}{OT1}{pzc}{m}{n}

\DeclareMathOperator*{\argmax}{arg\,max}

\newtheorem{thm}{Theorem}

\newcommand{\cD}{\mathcal{D}}
\newcommand{\cA}{\mathcal{A}}

\newcommand{\cR}{\mathcal{R}}
\newcommand{\cO}{\mathcal{O}}

\newcommand{\cS}{\mathcal{S}}

\newcommand{\cB}{\mathcal{B}}
\newcommand{\cU}{\mathcal{U}}
\newcommand{\cH}{\mathcal{H}}

\newcommand{\norm}[1]{\left\|#1\right\|_2}

\graphicspath{{figures/}}

\mathchardef\mhyphen="2D
\newcommand{\cascadehybrid}{\sf CascadeHybrid}

\newcommand{\cascadelinucb}{\sf CascadeLinUCB}
\newcommand{\cascadelsb}{\sf CascadeLSB}
\newcommand{\cascadelinucbfull}{\sf CascadeLinUCBFull}
\newcommand{\cascadelsbfull}{\sf CascadeLSBFull}

\newcommand{\lsbgreedy}{\sf LSBGreedy}
\newcommand{\ccucb}{\sf C^2UCB}
\newcommand{\recurrank}{\sf RecuRank}
\newcommand{\myparagraph}[1]{\medskip\noindent\textbf{#1.}}

\acrodef{IR}{information retrieval}
\acrodef{OLTR}{Online learning to rank}
\acrodef{LTR}{learning to rank}
\acrodef{CM}{cascade model}
\acrodef{CB}{cascading bandits}
\acrodef{MAB}{multi-armed bandits}
\acrodef{UCB}{upper confidence bound}
\acrodef{CHB}{cascade hybrid bandits}

\newcommand{\vOt}{\myvec{O}_t}

\allowdisplaybreaks

%% file: authors.tex
\author{Chang Li}
\orcid{1234-5678-9012}
\affiliation{%
	\institution{University of Amsterdam}
}
\email{c.li@uva.nl}

\author{Haoyun Feng}
\orcid{1234-5678-9012}
\affiliation{%
	\institution{Bloomberg}
}
\email{hfeng19@bloomberg.net}

\author{Maarten de Rijke}
\orcid{1234-5678-9012}
\affiliation{%
	\institution{University of Amsterdam}
}
\affiliation{%
	\institution{Ahold Delhaize}
}
\email{m.derijke@uva.nl}

%% file: sections/1-Introduction.tex

\section{Introduction}
\label{sec:introduction}

Ranking is at the heart of modern interactive systems, such as recommender and search systems. 
Learning to rank~(\acs{LTR})\acused{LTR} addresses the ranking problem in such systems by using machine learning approaches~\citep{liu2009learning}. 
Traditionally, \ac{LTR} has been studied in an offline fashion, in which human labeled data is required~\citep{liu2009learning}. 
Human labeled data is expensive to obtain, cannot capture future changes in user preferences, and may not well align with user needs~\citep{hofmann-balancing-2011}.
To circumvent these limitations, recent work has shifted to learning directly from users' interaction feedback, e.g., clicks~\citep{hofmann2013reusing,zoghi-click-2016,jagerman2019comparison}. 

User feedback is abundantly available in interactive systems and is a valuable source for training online \ac{LTR} algorithms~\citep{grotov-2016-online}. 
When designing an algorithm to learn from this source, three challenges need to be addressed: 
\begin{enumerate*}
	\item The learning algorithm should address position bias (the phenomenon that higher ranked items are more likely be observed than lower ranked items); 
	\item The learning algorithm should infer item relevance from user feedback and recommend lists containing relevant items (relevance ranking);
	\item The recommended list should contain no redundant items and cover a broad range of topics (result diversification).
\end{enumerate*}

To address the position bias, a common approach is to make assumptions on the user's click behavior and model the behavior using a click model~\cite{chuklin2015click}.
The cascade model~(\acs{CM})\acused{CM}~\citep{craswell08experimental} is a simple but effective click model to explain user behavior.
It makes the so-called \emph{cascade assumption}, which assumes that a user browses the list from the first ranked item to the last one and clicks on the first attractive item and then stops browsing. 
The clicked item is considered to be positive, items before the click are treated as negative and items after the click will be ignored. 
Previous work has shown that the cascade assumption can explain the position bias effectively and several algorithms have been proposed under this assumption~\citep{kveton15cascading,zong16cascading,Hiranandani2019CascadingLS,li-2019-cascading}. 

In online \ac{LTR}, the implicit signal that is inferred from user interactions is noisy~\citep{hofmann-balancing-2011}. 
If the learning algorithm only learns from these signals, it may reach a suboptimal solution where the optimal ranking is ignored simply because it is never exposed to users.
This problem can be tackled by exploring new solutions, where the learning algorithm displays some potentially ``good'' rankings to users and obtains more signals. 
This behavior is called \emph{exploration}. 
However, exploration may hurt the user experience. 
Thus, learning algorithms face an exploration vs.\ exploitation dilemma. 
Multi-armed bandit~(\acs{MAB}) \acused{MAB}~\citep{auer02finitetime,lattimore2018bandit} algorithms are commonly used to address this dilemma. 
Along this line, multiple algorithms have been proposed~\citep{li2010contextual,hofmann2013reusing,oosterhuis-differentiable-2018,li2019bubblerank}. 
They all address the dilemma in elegant ways and aim at recommending the top-$K$ most relevant items to users. 
However, only recommending the most relevant items may result in a list with redundant items, which diminishes the utility of the list and decreases user satisfaction~\citep{Agrawal2009-diversifying,yue-2011-linear}. 

The submodular coverage model~\citep{nemhauser1978analysis} can capture the pattern of diminishing utility and has been used in online \ac{LTR} for diversified ranking.
One assumption in this line of work is that items can be represented by a set of topics.\footnote{In general, each topic may only capture a tiny aspect of the information of an item, e.g., a single phrase of a news title or a singer of a song~\citep{yue-2011-linear,Agrawal2009-diversifying}.} 
The task, then, is to recommend a list of items that ensures a maximal coverage of topics. 
\citet{yue-2011-linear} develop an online feature-based diverse \ac{LTR} algorithm by optimizing submodular utility models~\citep{yue-2011-linear}. 
\citet{Hiranandani2019CascadingLS} improve online diverse \ac{LTR} by bringing the cascading assumption into the objective function. 
However, we argue that not all features that are used in a \ac{LTR} setting can be represented by topics~\citep{liu2009learning}.
Previous online diverse \ac{LTR} algorithms tend to ignore the relevance of individual items and may recommend a diversified list with less relevant items. 

In this work, we address the aforementioned challenges and make four contributions: 
\begin{enumerate}
	\item We focus on a novel online \ac{LTR} setting that targets both relevance ranking and result diversification.
	We formulate it as a \acf{CHB} problem, where the goal is to select $K$ items from a large candidate set that maximize the utility of the ranked list (\cref{sec:problem formulation}).
	\item We propose $\cascadehybrid$, which utilizes a hybrid model, to solve this problem~(\cref{sec:cascadehybrid}).  
	\item We evaluate $\cascadehybrid$ on two real-world recommendation datasets: MovieLens and Yahoo and show that\break $\cascadehybrid$ outperforms state-of-the-art baselines (\cref{sec:experiments}). 
	\item We theoretically analyze the performance of $\cascadehybrid$ and provide guarantees on its proper behavior; moreover, we are the first to show that the regret bounds on feature-based ranking algorithms with the cascade assumption are linear in $\sqrt{K}$. 
\end{enumerate}

\noindent%
The rest of the paper is organized as follows. 
We recapitulate the background knowledge in Section~\ref{sec:background}. 
In Section~\ref{sec:algorithm}, we formulate the learning problem and propose our $\cascadehybrid$ algorithm that optimizes both item relevance and list diversity. 
Section~\ref{sec:experiments} contains our empirical evaluations of $\cascadehybrid$, comparing it with several state-of-the-art baselines. 
An analysis of the upper bound on the $n$-step performance of $\cascadehybrid$ is presented in Section~\ref{sec:analysis}. 
In Section~\ref{sec:related work}, we review related work. 
Conclusions are formulated in Section~\ref{sec:conclusion}.    

%% file: sections/2-Background.tex

\section{Background}
\label{sec:background}

In this section, we recapitulate the \acf{CM}, \ac{CB}, and the submodular coverage model.
Throughout the paper, we consider the ranking problem of $L$ candidate items and $K$ positions with $K \leq L$. 
We denote $\{1, \ldots, n\}$ by $[n]$ and for the collection of items we write $\cD = [L]$.  
A ranked list contains $K \leq L$ items and is denoted by $\cR \in \Pi_K(\cD)$, where $\Pi_K{(\cD)}$ is the set of all permutations of $K$ distinct items from the collection $\cD$. 
The item at the $k$-th position of the list is denoted by $\cR(k)$ and, if $\cR$ contains an item $i$, the position of this item in $\cR$ is denoted by $\cR^{-1}(i)$. 
All vectors are column vectors.
We use bold font to indicate a vector and bold font with a capital letter to indicate a matrix.
We write $\vI_{d}$ to denote the $d\times d$ identity matrix and $\mathbf{0}_{d\times m}$ the $d\times m$ zero matrix.

\subsection{Cascade model}
\label{sec:cascade models}
Click models have been widely used to interpret user's interactive click behavior;  cf.~\citep{chuklin2015click}.  
Briefly, a user is shown a ranked list $\cR$, and then browses the list and leaves click feedback.  
Every click model makes unique assumptions and models a type of user interaction behavior. 
In this paper, we consider a simple but widely used click model, the \emph{cascade model}~\citep{craswell08experimental,kveton15cascading,li-2019-cascading}, which makes the cascade assumption about user behavior. 
Under the cascade assumption, a user browses a ranked list $\cR$ from the first item to the last one by one and clicks the first attractive item. 
After the click, the user stops browsing the remaining items. 
A click on an examined item $\cR(i)$ can be modeled as a Bernoulli random variable with a probability of $\alpha(\cR(i))$, which is also called the \emph{attraction probability}.
Here, the \acf{CM} assumes that each item attracts the user independent of other items in $\cR$. 
Thus, the \ac{CM} is parametrized by a set of attraction probabilities $\valpha \in [0, 1]^L$. 
The examination probability of item $\cR(i)$ is $1$ if $i=1$, otherwise $1-\prod_{j=1}^{i-1}(1-\alpha(\cR(j)))$. 

With the \ac{CM}, we translate the implicit feedback to training labels as follows: 
Given a ranked list, items ranked below the clicked item are ignored since none of them are browsed. 
Items ranked above the clicked item are negative samples and the clicked item is the positive sample. 
If no item is clicked, we know that all items are browsed but not clicked. 
Thus, all of them are negative samples. 

The vanilla \ac{CM} is only able to capture the first click in a session, and there are various extensions of \ac{CM} to model multi-click scenarios; cf.~\citep{chuklin2015click}. 
However,  we still focus on the \ac{CM}, because it has been shown in multiple publications that the \ac{CM} achieves good performance in both online and offline setups~\citep{chuklin2015click,kveton15cascading,li2019bubblerank}.

\subsection{Cascading bandits}
\label{sec:cascading click bandits}

Cascading bandits (\acs{CB})\acused{CB} are a type of online variant of the \ac{CM}~\citep{kveton15cascading}. 
A \ac{CB} is represented by a tuple $(\cD, K, P)$, where $P$ is a binary distribution over  $\{0, 1\}^L$.  
The learning agent interacts with the \ac{CB} and learns from the feedback. 
At each step $t$, the agent generates a ranked list $\cR_t \in \Pi_K(\cD)$ depending on observations in the previous $t-1$ steps and shows it to the user. 
The user browses the list with cascading behavior and leaves click feedback. 
Since the \ac{CM} accepts at most one click, we write $c_t \in [K+1]$ as the click indicator,  where $c_t$ indicates the position of the click and $c_t = K+1$ indicates no click. 
Let $A_t \in \{0, 1\}^L$ be the attraction indicator, where $A_t$ is drawn from $P$ and $A_t(\cR_t(i))=1$ indicates that item $\cR_t(i)$ attracts the user at step $t$. 
The number of clicks at step $t$ is considered as the reward and computed as follows: 
\begin{equation}
\label{eq:reward}
r(\cR_t, A_t) = 1 - \prod_{i=1}^{K} (1-A_t(\cR_t(i))).
\end{equation}
Then, we assume that the attraction indicators of items are distributed independently as Bernoulli variables: 
\begin{equation}
\label{eq:bernoulli}
	P(A) = \prod_{i \in \cD} P_{\alpha(i)}(A(i)),
\end{equation}
where $P_{\alpha(i)}(\cdot)$ is the Bernoulli distribution with mean $\alpha(i)$.
The expected number of clicks at step $t$ is computed as $\expect{r(\cR_t, A_t)} = r(\cR_t, \valpha)$. 
The goal of the agent is to maximize the expected number of clicks in $n$ steps or minimize the expected $n$-step regret: 
\begin{equation}
\label{eq:regret}
	R(n) = \sum_{t=1}^{n}\expect{  \max_{\cR \in \Pi_K(\cD)} r(\cR, \valpha) - r(\cR_t, A_t) }.
\end{equation}
\ac{CB} has several variants depending on assumptions on the attraction probability $\valpha$.
Briefly, cascade linear bandits~\citep{zong16cascading} assume that an item $a$ is represented by a feature vector $\vz_a \in \real^{m}$ and that the attraction probability of an item $a$ to a user is a linear combination of features: $ \alpha(a)\approx \vz_a^T \beta^*$, where $\beta^* \in \real^m$ is an unknown parameter. 
With this assumption, the attraction probability of an item is independent of other items in the list, and  this assumption is used in relevance ranking problems. 
$\cascadelinucb$ has been proposed to solve this problem. 
For other problems, \citet{Hiranandani2019CascadingLS} assume the attraction probability to be submodular, and propose $\cascadelsb$ to solve result diversification. 

\subsection{Submodular coverage model}
\label{sec:submodular coverage model}

Before we recapitulate the submodular function, we introduce two properties of a diversified ranking. 
Different from the relevance ranking, in a diversified ranking, the utility of an item depends on other items in the list. 
Suppose we focus on news recommendation.
Items that we want to rank are news itms, and each news item covers a set of topics, e.g., weather, sports, politics, a celebrity, etc. 
We want to recommend a list that covers a broad range of topics. 
Intuitively, adding a news item to a list does not decrease the number of topics that are covered by the list, but adding a news item to a list that covers highly overlapping topics might not bring much extra benefit to the list. 
The first property can be thought of as a monotonicity property, and the second one is the notion of diminishing gain in the utility. 
They can be captured by the submodular function~\cite{yue-2011-linear}. 

We introduce two properties of submodular functions. 
Let $g(\cdot)$ be a set function, which maps a set to a real value. 
We say that $g(\cdot)$ is monotone and submodular if given two item sets $\cA$ and $\cB$, where  $\cB \subseteq \cA$, and an item $a$, $g(\cdot)$ has the following two properties:  
\begin{equation*}
\begin{aligned}
&\emph{monotonicity}: g(\cA \cup \{a\}) \geq g(\cA);  \\
&\emph{submodularity}: g(\cB \cup \{a\}) - g(\cB) \geq	g(\cA \cup \{a\}) - g(\cA). 
\end{aligned}
\end{equation*}

\noindent
In other words, the gain in utility of adding an item $a$ to a subset of $\cA$ is larger than or equal to that of adding an item to $\cA$, and adding an item $a$ to $\cA$ does not decreases the utility. 
Monotonicity and submodularity together provide a natural framework to capture the properties of a diversified ranking.
The shrewd reader may notice that a linear function is a special case of submodular functions, where only the inequalities in \emph{monotonicity} and \emph{submodularity} hold. 
However, as discussed above, the linear model assumes that the attraction probability of an item is independent of other items: it cannot capture the diminishing gain in the result diversification. 
In the rest of this section, we introduce the \emph{probabilistic coverage model}, which 
is a widely used submodular function for result diversification~\citep{Agrawal2009-diversifying,yue-2011-linear,Qin2013-promoting,Ashkan2015-optimal,Hiranandani2019CascadingLS}.

Suppose that an item $a\in\cD$ is represented by a $d$-dimensional vector $\vx_a \in [0, 1]^d$. 
Each entry of the vector $\vx_a(j)$ describes the probability of item $a$ covering topic $j$.
Given a list $\cA$, the probability of $\cA$ covering topic $j$ is 
\begin{equation}
	\label{eq:topic coverage}
g_j(\cA) = 1-\prod_{a \in \cA} (1-\vx_a(j)). 
\end{equation}
The gain in topic coverage of adding an item $a$ to $\cA$ is:
\begin{equation}
\label{eq:topic gain}
\Delta(a\mid \cA) = (\Delta_1(a\mid \cA), \ldots, \Delta_d(a\mid \cA)), 
\end{equation}
where $\Delta_j(a\mid \cA) = g_j(\cA\cup\{a\}) - g_j(\cA)$. 
With this model, the attraction probability of the $i$-th item in a ranked list $\cR$ is defined as: 
\begin{equation}
	\label{eq:topic coverage attration probability}
	\alpha(\cR(i)) = \vomega_{\cR(i)}^T \vtheta^*, 
\end{equation}
where $\vomega_{\cR(i)} = \Delta(\cR(i) \mid (\cR(1), \ldots, \cR(i-1)))$ and $\vtheta^*$ is the unknown user preference to different topics~\citep{Hiranandani2019CascadingLS}. 
In \cref{eq:topic coverage attration probability},  the attraction probability of an item depends on the items ranked above it; $\alpha(\cR(i))$ is small if $\cR(i)$ covers similar topics as higher ranked items.  
$\cascadelsb$~\citep{Hiranandani2019CascadingLS} has been proposed to solve cascading bandits with this type of attraction probability and aims at building diverse ranked lists. 

%% file: sections/3-Algorithm.tex

\section{Algorithm}
\label{sec:algorithm}
In this section, we first formulate our online learning to rank problem, and then propose $\cascadehybrid$ to solve it. 

\subsection{Problem formulation}
\label{sec:problem formulation}
We study a variant of cascading bandits, where the attraction probability of 
an item in a ranked list depends on two aspects: item relevance and item novelty. 
Item relevance is independent of other items in the list.
Novelty of an item depends on the topics covered by higher ranked items;
a novel item brings a large gain in the topic coverage of the list, i.e., a large value in \cref{eq:topic coverage attration probability}.
Thus, given a ranked list $\cR$, the attraction probability of item $\cR(i)$ is defined as follows: 
\begin{equation}
	\label{eq:hybridy attraction probability}
	\alpha(\cR(i)) =  \vz_{\cR(i)}^T\vbeta^* +  \vomega_{\cR(i)}^T\vtheta^*, 
\end{equation} 
where $\vomega_{\cR(i)}= \Delta(\cR(i) \mid (\cR(1), \ldots, \cR(i-1))$ is the topic coverage gain discussed in \cref{sec:submodular coverage model}, and $\vz_{\cR(i)} \in \real^m$ is the relevance feature, $\vtheta^* \in \real^d$ and $\vbeta^* \in \real^m$ are two unknown parameters that characterize the user preference. 
In other words, the attraction probability is a hybrid of a modular (linear) function parameterized by $\vbeta^*$ and a submodular function parameterized by $\vtheta^*$. 

Now, we define our learning problem, \acf{CHB}, as a tuple $(\cD, \vtheta^*, \vbeta^*,  K)$. 
Here, $\cD = [L]$ is the item candidate set and each item $a$ can be represented by a feature vector $[\vx_a^T, \vz_a^T]^T$, where $\vx_a \in [0, 1]^d$ is the topic coverage of item $a$ discussed in \cref{sec:submodular coverage model}.
$K$ is the number of positions.
The action space for the problem are all permutations of $K$ individual items from $\cD$, $\Pi_K(\cD)$. 
The reward of an action at step $t$ is the number of clicks, defined in \cref{eq:reward}. 
Together with \cref{eq:reward,eq:hybridy attraction probability,eq:bernoulli},  the expectation of reward at step $t$ is computed as follows:
\begin{equation}
\expect{r(\cR_t, A_t)} = 1-\prod_{a \in \cR_t} (1- \vz_a^T\vbeta^* -  \vomega_a^T\vtheta^*). 
\end{equation}
In the rest of the paper, we write $r(\cR_t) = \expect{r(\cR_t, A_t)}$ for short. 
And the goal of the learning agent is to maximize the  reward or, equivalently, to minimize the  $n$-step regret defined as follow: 
\begin{equation}
\label{eq:our regret}
R(n) = \sum_{t=1}^{n} \left[\max_{\cR\in \Pi_K(\cD)}r(\cR) - r(\cR_t)\right].
\end{equation}

\noindent%
The previously proposed  $\cascadelinucb$~\citep{zong16cascading} cannot solve \ac{CHB} since it only handles the linear part of the attraction probability. 
$\cascadelsb$~\citep{Hiranandani2019CascadingLS} cannot solve  \ac{CHB}, either, because it only uses one submodular function and handles the submodular part of the attraction probability. 
Thus, we need to extend the previous models or, in other words, propose a new hybrid model that can handle both linear and submodular properties in the attraction probability.  

\subsection{Competing with a greedy benchmark}
\label{eq: optimal list}

Finding the optimal set that maximizes the utility of a submodular function is an NP-hard problem~\citep{nemhauser1978analysis}. 
In our setup, the attraction probability of each item also depends on the order in the list. 
To the best of our knowledge, we cannot find the optimal ranking 
\begin{equation}
\cR^* = \argmax_{\cR\in \Pi_K(\cD)}r(\cR)
\end{equation} 
efficiently. 
Thus, we compete with a greedy benchmark that approximates the optimal ranking $\cR^*$. 
The greedy benchmark chooses the items that have the highest attraction probability given the higher ranked items:  for any positions $k \in [K]$, 
\begin{equation}
\tilde{\cR}(k) = \argmax\limits_{a \in \cD\backslash \{\tilde{\cR}(1),\ldots, \tilde{\cR}(k-1)\}} \vz_a^T\vbeta^* +  \vomega_a^T\vtheta^*,
\end{equation}
where $\tilde{\cR}(k)$ is the ranked list generated by the benchmark. 

This greedy benchmark has been used in previous literature~\citep{yue-2011-linear,Hiranandani2019CascadingLS}. 
As shown by~\citet{Hiranandani2019CascadingLS}, in the \ac{CM}, the greedy benchmark is at least a $\eta$-approximation of $\cR^*$.
That is, 
$	r(\tilde{\cR}) \geq \eta r(\cR^*)$
where $\eta = (1-\frac{1}{e}) \max\{\frac{1}{K}, 1-\frac{K-1}{2}\alpha_{max}\}$ with $\alpha_{max} = \max_{a \in \cD} \vz_a^T\vbeta^* +  \vx_a^T\vtheta^*$.
In the rest of the paper, we focus on competing with this greedy benchmark.

\begin{algorithm}[t]
	\algrenewcommand{\algorithmiccomment}[1]{// #1}
	\algnewcommand\algorithmicinput{\textbf{Input:}}
	\algnewcommand\Input{\item[\algorithmicinput]}
	\algnewcommand\algorithmicoutput{\textbf{Output:}}
	\algnewcommand\Output{\item[\algorithmicoutput]}
	\begin{algorithmic}[1]
		\Input $\gamma$
		\State 	\Comment{Initialization} 
		\State \label{alg:initialization}
		$\vH_1 \leftarrow \vI_{d}, 
		\vu_1 \leftarrow \mathbf{0}_{d}, 
		\vM_1 \leftarrow \vI_{m}, 
		\vy_1 \leftarrow \mathbf{0}_{m}, 
		\vB_1 = \mathbf{0}_{d\times m} $
		
		\For{$t = 1, 2, \ldots, n$}
		\State \Comment{Estimate parameters}
		\State \label{alg:estimate parameters}
		$\hat{\vtheta}_{t} \leftarrow \vH_{t}^{-1}\vu_{t}$, 
		$\hat{\vbeta}_{t}\leftarrow \vM_{t}^{-1}(\vy_{t} - \vB_{t}^T  \vH_{t}^{-1}\vu_{t})$
		
		\State \Comment{Build ranked list}
		\State \label{alg:start building}
		$\cS_0 \leftarrow \emptyset$
		\For{$k = 1, 2, \ldots K$}
		\For{$a \in \cD \setminus \cS_{k-1}$} \label{alg:begin estimate}
		\State $\vomega_a \leftarrow \Delta(\vx_a | \cS_{k-1})$
		\Comment{Recalculate the topic coverage gain. }
		\State $\mu_{a} \leftarrow$  \cref{eq:ucb}  \Comment{Compute UCBs.}
		\EndFor \label{alg:end estimate}
		
		\State \label{alg:argmax}
		$a_k^t \leftarrow \argmax\limits_{a\in \cD \setminus \cS_{k-1}} \mu_{a} $
		\State $\cS_k \leftarrow \cS_{k-1} + {a_k^t}$
		\EndFor
		\State \label{alg:end building}
		$\cR_t = (a_1^t, \ldots, a_K^t)$ \Comment Ranked list
		
		\State \label{alg:display ranking}
		Display $\cR_t$ and observe click feedback $c_t \in [K+1]$
		\State $k_t \leftarrow \min(K, c_t)$
		\State \Comment{Update statistics}
		\State $\vH_{t} \leftarrow \vH_{t} + \vB_{t} \vM_{t}^{-1}\vB_{t}^T$, $\vu_{t}\leftarrow \vu_{t} + \vB_{t}\vM_{t}^{-1}\vy_{t}$
		\For{$a \in \cR_t(1:k_t)$}
		\State $\vM_{t+1} \leftarrow \vM_{t} + \vz_a \vz_a^T $, 
		$\vB_{t+1} \leftarrow \vB_{t} + \vomega_a\vz_a^T$, 
		$\vH_{t} \leftarrow \vH_{t} +  \vomega_a \vomega_a^T$			
		\EndFor
		\If{$c_t \leq K$}
		\State $\vy_{t+1} \leftarrow \vy_{t} + \vz_{\cR_t(c_t)}$, 
		$\vu_{t} \leftarrow \vu_{t} + \vomega_{\cR_t(c_t)}$
		\EndIf
		\State \label{alg:end click collection} \hspace{-0.11in}$\vH_{t+1} \leftarrow \vH_{t} - \vB_{t+1} \vM_{t+1}^{-1}\vB_{t+1}^T$, $\vu_{t+1} \leftarrow \vu_{t} - \vB_{t+1}\vM_{t+1}^{-1}\vy_{t+1}$
		\EndFor
	\end{algorithmic}
	\caption{$\cascadehybrid$}
	\label{alg:cascadehybrid}
\end{algorithm}
 
\subsection{\textsf{\textbf{CascadeHybrid}}}
 \label{sec:cascadehybrid}
We propose $\cascadehybrid$ to solve the \ac{CHB}.
As the name suggests, the algorithm is a hybrid of a linear function and a submodular function. 
The linear function is used to capture item relevance and the submodular function to capture diversity in topics. 
$\cascadehybrid$ has access to item features, $[\vx_a^T, \vz_a^T]^T$, and uses the probabilistic coverage model to compute the gains in topic coverage. 
The user preferences $\vtheta^*$ and $\vbeta^*$ are unknown to $\cascadehybrid$. 
They are estimated from interactions with users. 
The only tunable hyperparameter for $\cascadehybrid$ is $\gamma \in \real_+$, which controls exploration: a larger value of $\gamma$ means more exploration. 

The details of $\cascadehybrid$ are provided in \cref{alg:cascadehybrid}. 
At the beginning of each step $t$ (line~\ref{alg:estimate parameters}), $\cascadehybrid$ estimates the user preference as $\hat{\vtheta}_t$ and $\hat{\vbeta}_{t}$ based on the previous $t-1$ step observations. 
$\hat{\vtheta}_t$ and $\hat{\vbeta}_{t}$ can be viewed as maximum likelihood estimators on the rewards,\footnote{The derivation is based on matrix block-wise inversion. We omit the derivation since it is not a major contribution of this paper. 
} 
where $\vM_{t}, \vH_{t}, \vB_{t}$ and $\vy_t, \vu_t$ summarize the features and click feedback of all observed items in the previous $t-1$ steps.
Then, $\cascadehybrid$ builds the ranked list $\cR_t$, sequentially~(line~\ref{alg:start building}--\ref{alg:end building}). 
In particular, for each position $k$, we recalculate the topic coverage gain of each item (line~\ref{alg:start building}). 
The new gains are used to estimate the attraction probability of items. 
$\cascadehybrid$ makes an optimistic estimate of the attraction probability of each item (line~\ref{alg:begin estimate}--\ref{alg:end estimate}) and chooses the one with the highest  estimated attraction probability (line~\ref{alg:argmax}). 
This is known as the principle of  optimism in the face of uncertainty~\citep{auer02finitetime}, and the estimator for an item $a$ is called the \ac{UCB}: 
\begin{equation}
\label{eq:ucb}
	\mu_{a} =  \vomega_a^T \hat{\vtheta}_t + \vz_a^T \hat{\vbeta}_t  + \gamma\sqrt{s_{a}},  
\end{equation}
with 
\begin{equation}
\label{eq:cb}
\begin{aligned}
s_{a}  =&  \vomega_a^T\vH_{t}^{-1}\vomega_a - 2\vomega_a^T\vH_{t}^{-1}\vB_{t}\vM_{t}^{-1}\vz_a +  \vz_a \vM_{t}^{-1} \vz_a + \\
&\quad \vz_a \vM_{t}^{-1}\vB_{t}^T\vH_{t}^{-1}\vB_{t} \vM_{t}^{-1} \vz_a.
\end{aligned}
\end{equation}
Finally, $\cascadehybrid$ displays the ranked list $\cR_t$ to the user and collects click feedback (line~\ref{alg:display ranking}--\ref{alg:end click collection}). 
Since $\cascadehybrid$ only accepts one click, we use $c_t \in [K+1]$ to indicate the position of the click;\footnote{
	For multiple-click cases, we only consider the first click and keep the rest of $\cascadehybrid$ the same. }
$c_t = K+1$ indicates that no item in $\cR_t$ is clicked. 

\subsection{Computational complexity}
\label{eq:computational complexity}

The main computational cost of \cref{alg:cascadehybrid} is incurred by computing matrix inverses, which is cubic in the dimensions of the matrix. 
However, in practice, we can use the Woodbury matrix identity~\citep{golub-1996-matrix} to update $\vH_{t}^{-1}$ and $\vM_{t}^{-1}$ instead of $\vH_{t}$ and $\vM_{t}$, which is square in the dimensions of the matrix. 
Thus, computing the \ac{UCB} of each item is $O(m^2+d^2)$. 
As $\cascadehybrid$ greedily  chooses $K$ items out of $L$, the per-step computational complexity of $\cascadehybrid$ is $O(LK(m^2+d^2))$.

%% file: sections/4-Experiments.tex

\section{Experiments}
\label{sec:experiments}

This section starts with the experimental setup, where we first introduce the datasets, click simulator and baselines.
After that we report our experimental results.  

\subsection{Experimental setup}
\label{sec:experimental setup}

Off-policy evaluation~\citep{li2010contextual} is an approach to evaluate interaction algorithms without live experiments. 
However, in our problem, the action space is exponential in $K$, which is too large for  commonly used off-policy evaluation methods. 
As an alternative, we evaluate the $\cascadehybrid$ in a simulated interaction environment, where the simulator is built based on offline datasets. 
This is a commonly used evaluation setup in the literature~\citep{kveton15cascading,zong16cascading,Hiranandani2019CascadingLS}.

\myparagraph{Datasets}
We evaluate $\cascadehybrid$ on two datasets: MovieLens 20M~\citep{Harper:2015:movielense} and  Yahoo.\footnote{R2 - Yahoo! Music User Ratings of Songs with Artist, Album, and Genre Meta Information, v. 1.0~\url{https://webscope.sandbox.yahoo.com/catalog.php?datatype=r}} 
The MovieLens dataset contains $20$M ratings on $27$k movies by $138$k users, with $20$ genres.\footnote{In both datasets, one of the $20$ genres is called \emph{unknown}. } 
Each movie belongs to at least one genre. 
The Yahoo dataset contains  over $700$M ratings of $136$k songs given by $1.8$M users and genre attributes of each song; we consider the top level attribute, which has $20$ different genres; each song belongs to a single genre. 
All the ratings in the two datasets are on a $5$-point scale. 
All movies and songs are considered as items and genres are considered as topics.  

\myparagraph{Data preprocessing} 
We follow the data preprocessing approach in~\citep{zong16cascading,Hiranandani2019CascadingLS,li-2019-online}. 
First, we extract the $1$k most  active users and the $1$k most rated items. 
Let $\cU = [1000]$ be the user set, and $\cD = [1000]$ be the item set.  
Then, the ratings are mapped onto a binary scale: 
rating $5$ is converted to $1$ and others to $0$. 
After this mapping, in the MovieLens dataset, about $7\%$ of the user-item pairs get rating $1$, and, in the Yahoo dataset, about $11\%$ of user-item pairs get rating $1$.
Then, we use the matrix  $\vF \in \{0, 1\}^{|\cU|\times |\cD|}$ to capture the converted ratings and $\vG \in \{0, 1\}^{|\cD|\times d}$ to record the items and topics, where $d$ is the number of topics and each entry $\vG_{jk}=1$ indicates that item $j$ belongs to topic $k$.

\myparagraph{Click simulator}
In our experiments, the click simulator follows the cascade assumption, and considers both  item relevance and diversity of the list.
To design such a simulator, we combine the simulators used in \citep{li-2019-online} and \citep{Hiranandani2019CascadingLS}.
Because of the cascading assumption, we only need to define the way of computing attraction probabilities of items in a list. 

We first divide the users into training and test groups evenly, i.e., $\vF_{train}$ and $\vF_{test}$. 
The training group  is used to estimate features of items used by online algorithms, while the test group are used to define the click simulator. 
This is to mimic the real-world scenarios that online algorithms estimate user preferences without knowing the perfect topic coverage of items.
Then, we follow \citep{li-2019-online} to obtain the relevance part of the attraction probability, i.e., $\vz$ and $\vbeta^*$, and the process in \citep{Hiranandani2019CascadingLS} to get the topic coverage of items, $\vx$, and the user preferences on topics, $\vtheta^*$.

In particular,  the relevance features $\vz$ are obtained by conducting singular-value decomposition on $\vF_{train}$. 
We pick the  $10$ largest singular values and thus the dimension of relevance features is $m=10$. 
Then, we normalize each relevance feature by the transformation: $\vz_a \leftarrow \frac{\vz_a}{\norm{\vz_a}}$, where $\norm{\vz_a}$ is the L2 norm of $\vz_a$. 
The user preference $\vbeta^*$ is computed by solving the least square on $\vF_{test}$ and then $\vbeta^*$ is normalized by the same transformation. 
Note that $\forall a \in \cD: \vz_a^T \vbeta^*\in [0, 1]$,  since $ \norm{\vz_a}=1$ and $\norm{\vbeta^*} = 1$.

Then, we follow the process in \citep{Hiranandani2019CascadingLS}. 
If item $a$ belongs to topic $j$, we compute the topic coverage of item $a$ to topic $j$ as the quotient of the number of users rating item $a$ to be attractive to the number of users who rate at least one item in topic $j$ to be attractive: 
\begin{equation}
\label{eq:topic_features}
\vx_{a, j} = \frac{\sum_{u \in \cU} F_{u, a} G_{a, j}}	
{\sum_{u \in \cU} \mathds{1}\{\exists a' \in \cD: F_{u, a'}G_{a', j} >0 \}}. 
\end{equation}
Given user $u$, the preference for topic $j$ is computed as the number of items rated to be attractive in topic $j$ over the number of items in all topics rated by $u$ to be attractive: 
\begin{equation}
\label{eq:topic_preferences}
\vtheta_j^* = \frac{\sum_{a \in \cD} F_{u, a} G_{a, j}}{\sum_{j'\in[d]} \sum_{a' \in \cD} F_{u, a'} G_{a', j'}} .
\end{equation}
For some cases, we may have $\exists a:~ \sum_{j \in [d]} \vx_{a, j} >  1$ and thus $\vx_{a}^T \vtheta^* > 1$. 
However, given the high sparsity in our datasets,  we have $\vx_{a}^T \vtheta^* \in[0, 1]$ for all items during our experiments. 

Finally, we combine the two parts and obtain the attraction probability used in our click simulator. 
To simulate different types of user preferences, we introduce a trade-off parameter $\lambda \in [0, 1]$, which is unknown to online algorithms, and compute the attraction probability of the $i$th item in $\cR$ as follows:
\begin{equation}
\alpha(\cR(i)) = \lambda \vz_{\cR(i)}^T\vbeta^* +   (1-\lambda)  \vomega_{\cR(i)}^T\vtheta^*. 
\end{equation}
By changing the value of $\lambda$, we simulate different types of user preference: a larger value of $\lambda$ means that the user prefers items to be relevant; a smaller value of $\lambda$ means that the user prefers the topics in the ranked list to be diverse. 

\myparagraph{Baselines}
We compare $\cascadehybrid$ to two online algorithms, each of which has two configurations. 
In total, we have four baselines, namely 
$\cascadelinucb$ and $\cascadelinucbfull$~\citep{zong16cascading}, and $\cascadelsb$ and  $\cascadelsbfull$~\citep{Hiranandani2019CascadingLS}. 
The first two only consider relevance ranking. 
The differences are that 
$\cascadelinucb$ takes $\vz$ as the  features,  while $\cascadelinucbfull$ takes $\{\vx, \vz\}$ as the features. 
$\cascadelsb$ and  $\cascadelsbfull$ only consider the result diversification, where $\cascadelsb$ takes $\vx$ as features, while $\cascadelsbfull$ takes $\{\vx, \vz\}$ as features. 
$\cascadelinucbfull$ and $\cascadelinucb$  are expected to  perform well when $\lambda \to  1$, and that $\cascadelsb$ and $\cascadelsbfull$ perform well when $\lambda \to 0$. 
For all baselines, we set the exploration parameter $\gamma = 1$ and the learning rate to $1$. 
This parameter setup is used in \citep{yue-2011-linear}, which leads to better empirical performance. 
We also set $\gamma=1$ for $\cascadehybrid$. 

We report the cumulative regret, \cref{eq:our regret}, within $50$k steps, called \emph{$n$-step regret}.  
The $n$-step regret is commonly used to evaluate bandit algorithms~\cite{yue-2011-linear,zong16cascading,Hiranandani2019CascadingLS,li-2019-cascading}. 
In our setup, it measures the difference in number of received clicks between the oracle that knows the ideal $\beta^*$ and $\theta^*$ and the online algorithm, e.g., $\cascadehybrid$, in $n$ steps. 
The lower regret means the more clicks received by the algorithm. 
We conduct our experiments with $500$ users from the test group and $2$ repeats per user.
In total, the results are averaged over $1$k repeats. 
We also include the standard errors of our estimates. 
To show the impact of different factors on the performance of online \ac{LTR} algorithms, we choose $\lambda \in \{0.0, 0.1, \ldots, 1.0\}$, the number of positions $K\in\{5, 10, 15, 20\}$, and the number of topics $d \in \{5, 10, 15, 20\}$. 
For the number of topics $d$, we choose the topics with the maximum number of items.

\begin{figure*}[t]
	\centering
	\includegraphics{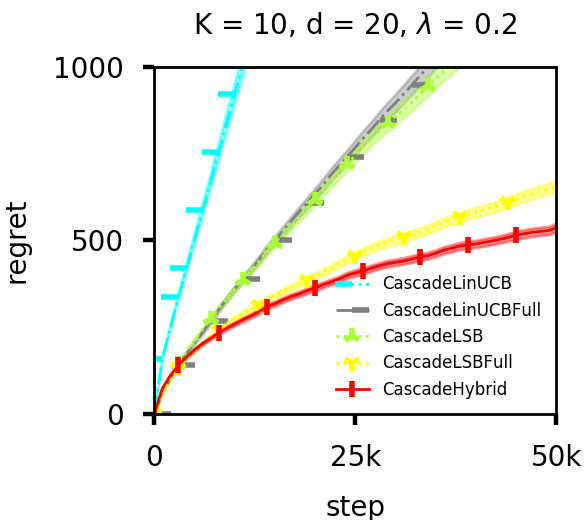}
	\includegraphics{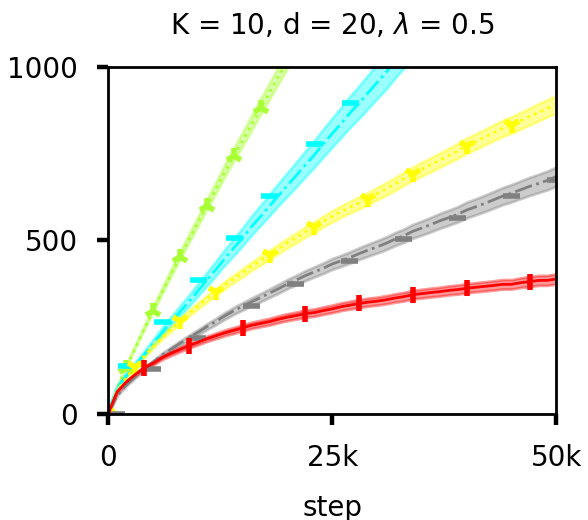}
	\includegraphics{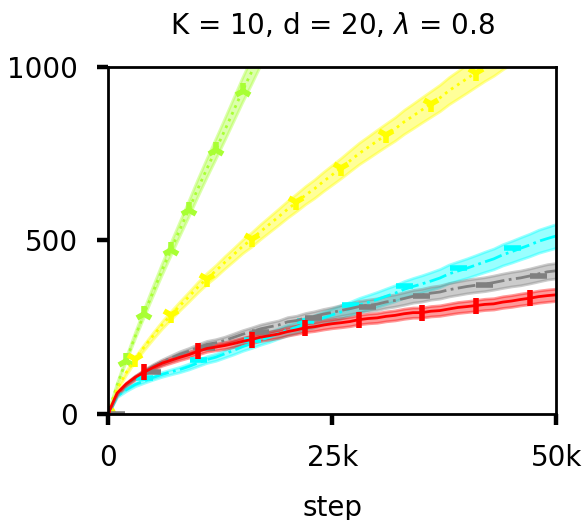}
	\includegraphics{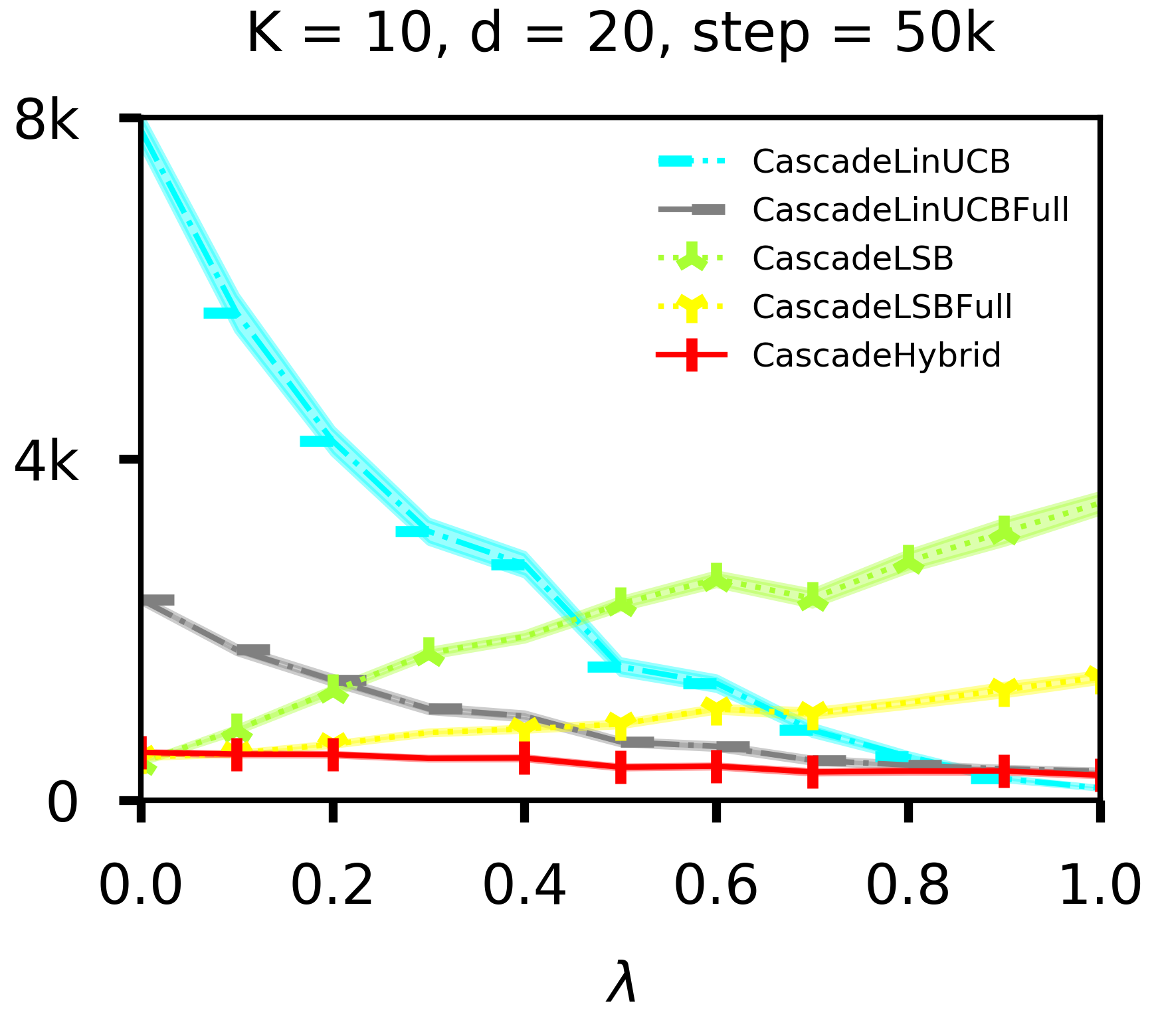}
	\includegraphics{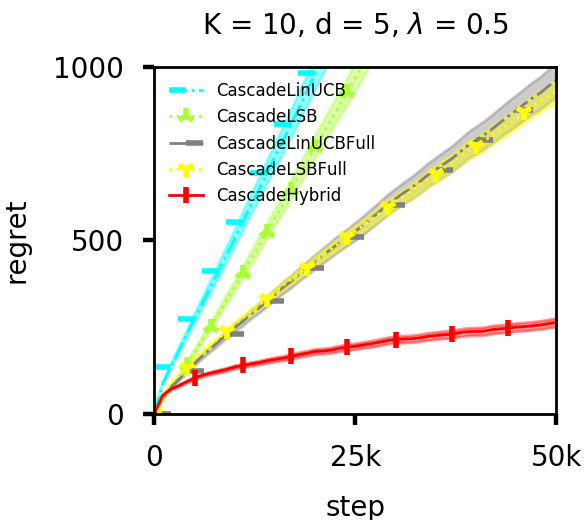}
	\includegraphics{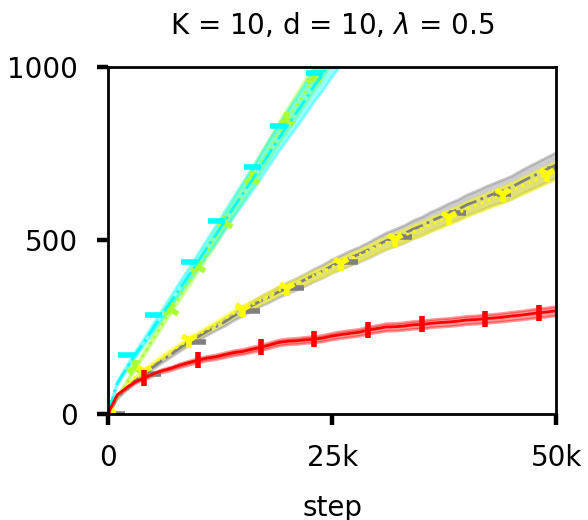}
	\includegraphics{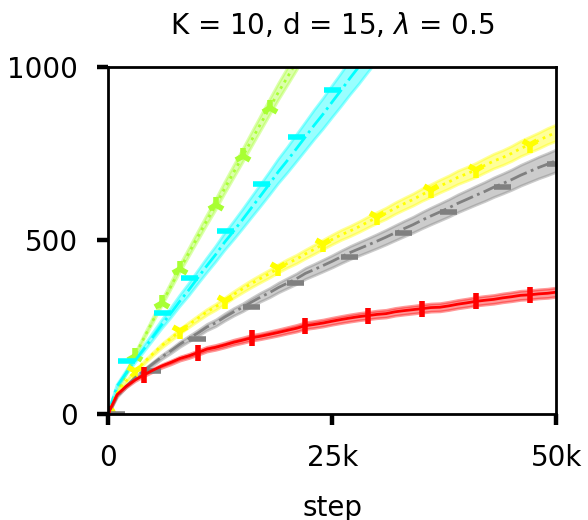}
	\includegraphics{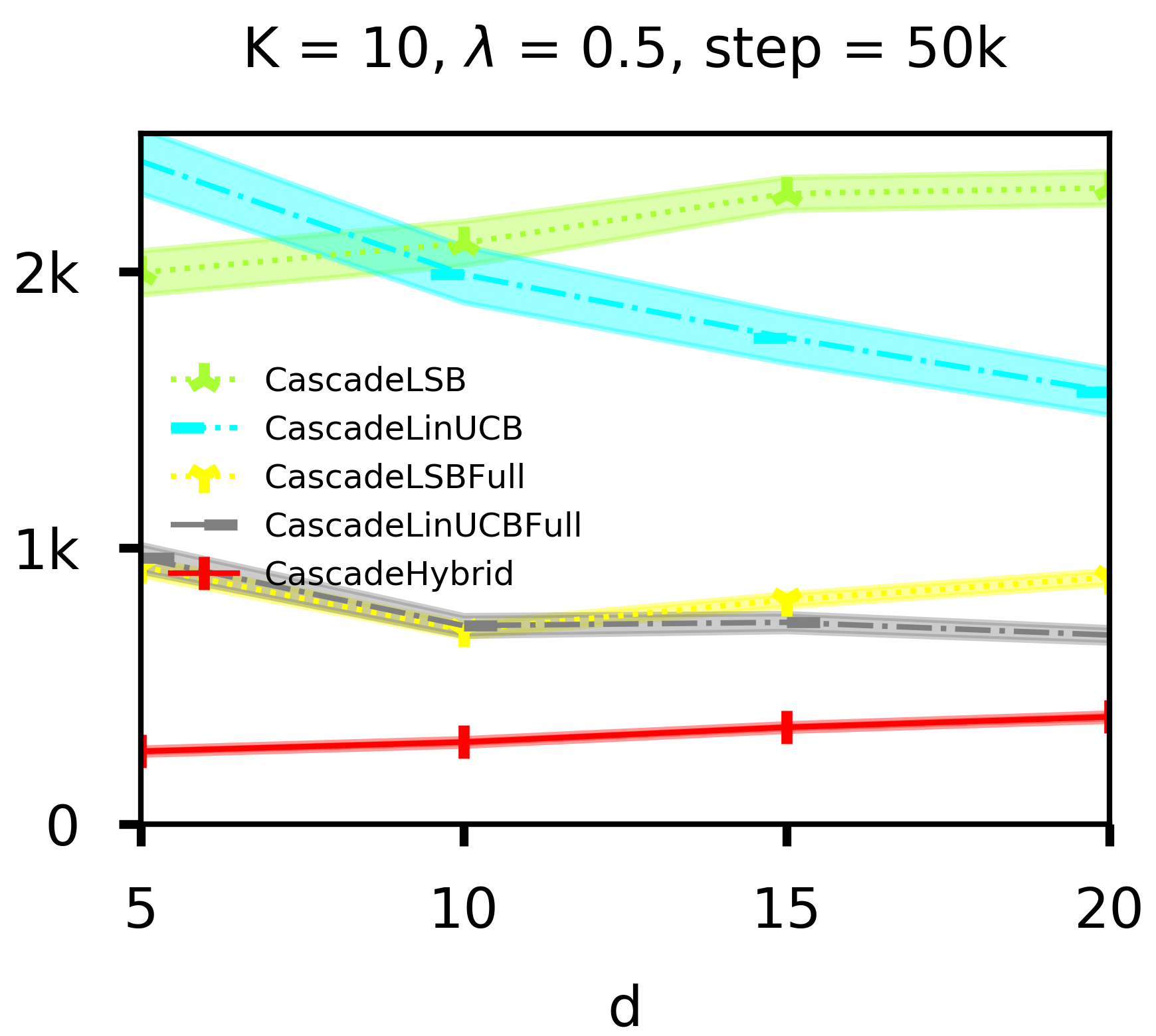}
	\includegraphics{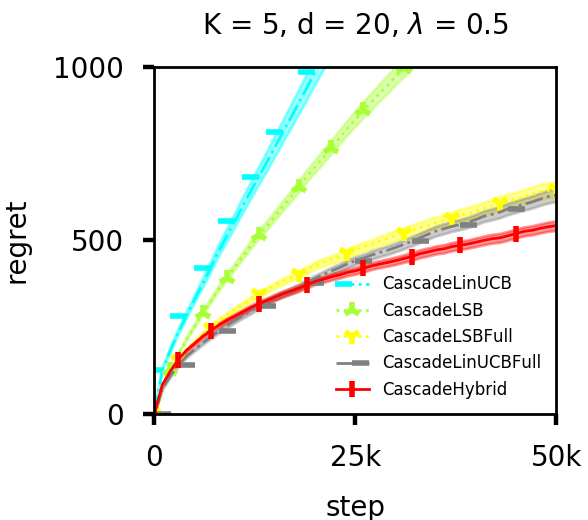}
	\includegraphics{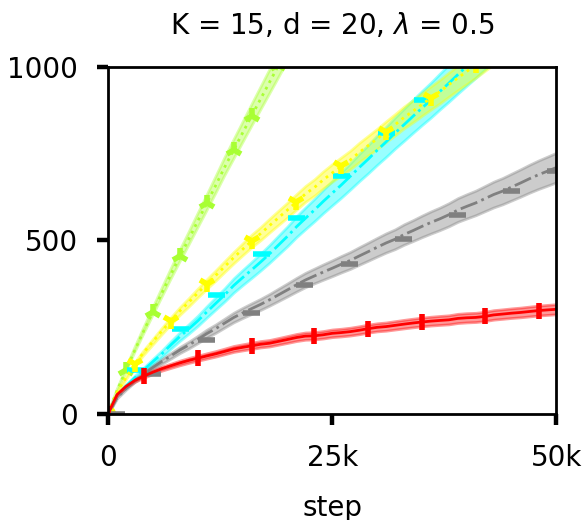}
	\includegraphics{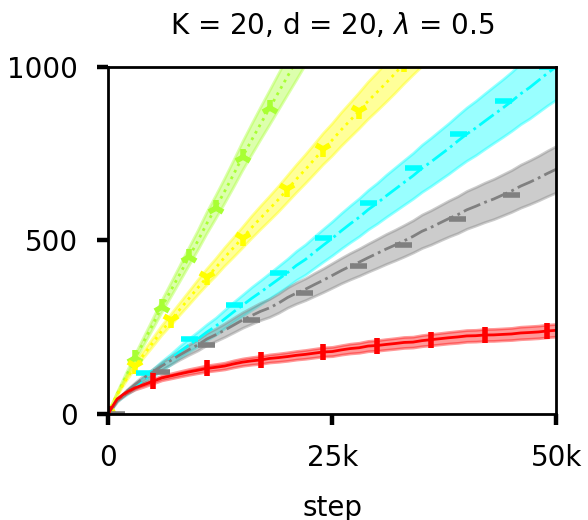}
	\includegraphics{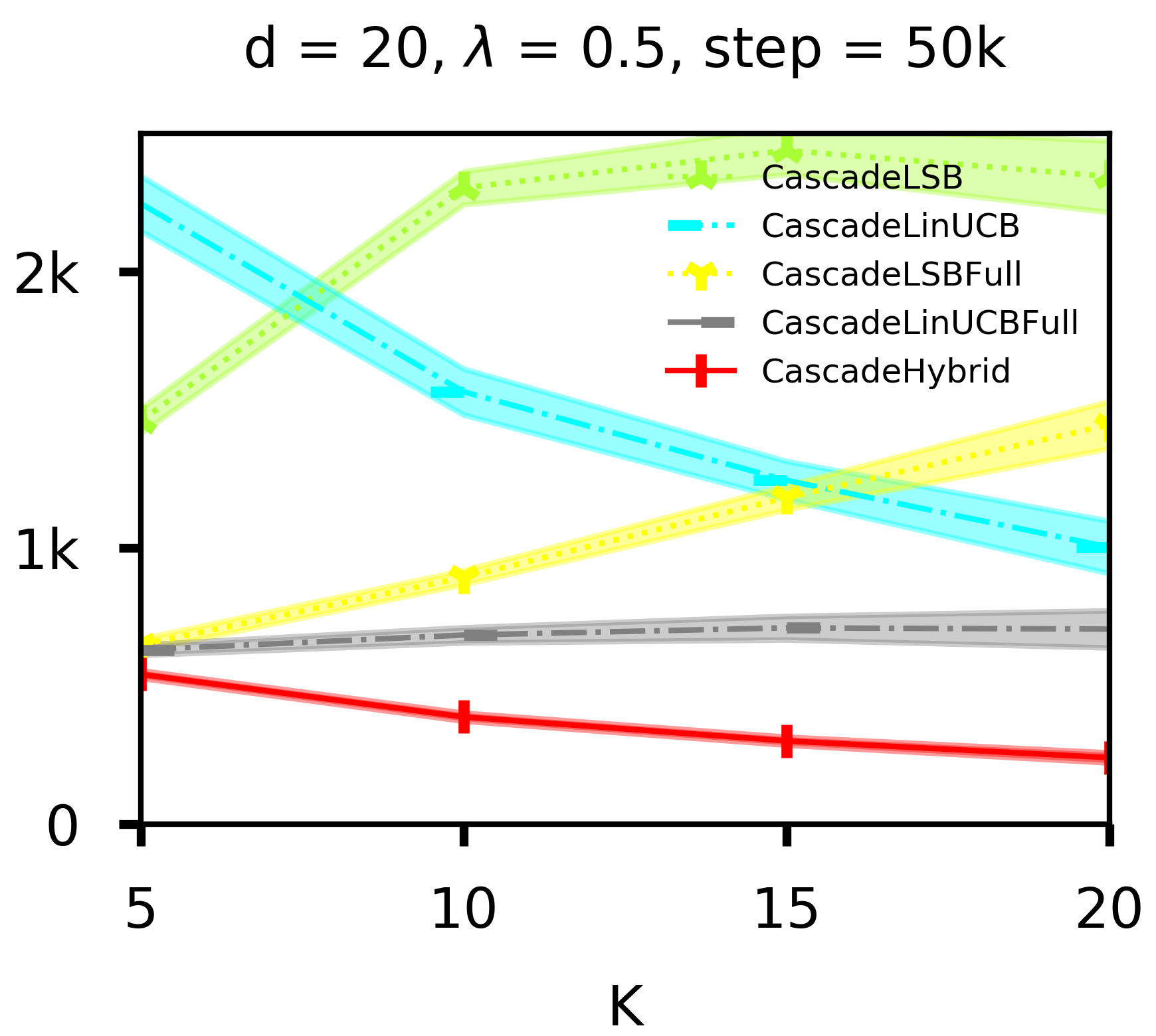}
	\caption{$n$-step regret on the MovieLens dataset. Results are averaged over $500$ users with $2$ repeats per user. 
		Lower regret means more clicks received by the algorithm during the online learning.
		Shaded areas are the standard errors of  estimates. }  
	\label{fig:movie}
\end{figure*}

\subsection{Experimental results}
\label{sec:experimental results}

\begin{figure*}[t]
	\centering
	\includegraphics{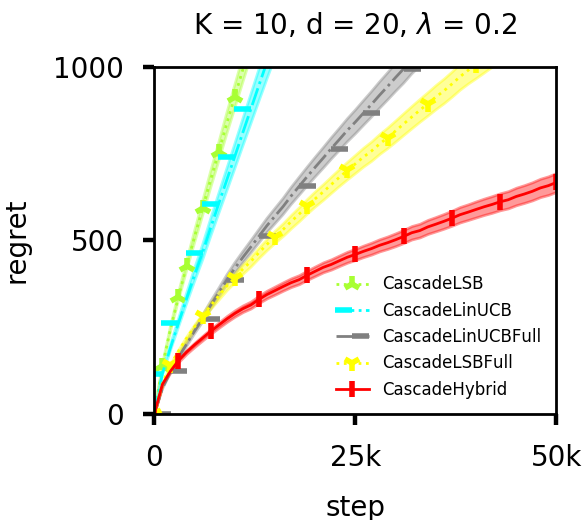}
	\includegraphics{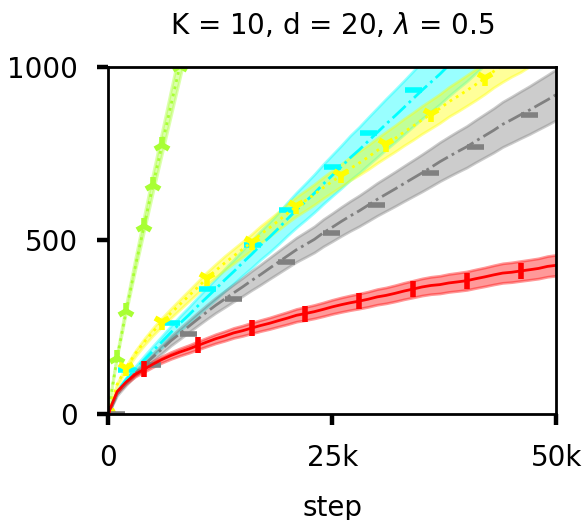}
	\includegraphics{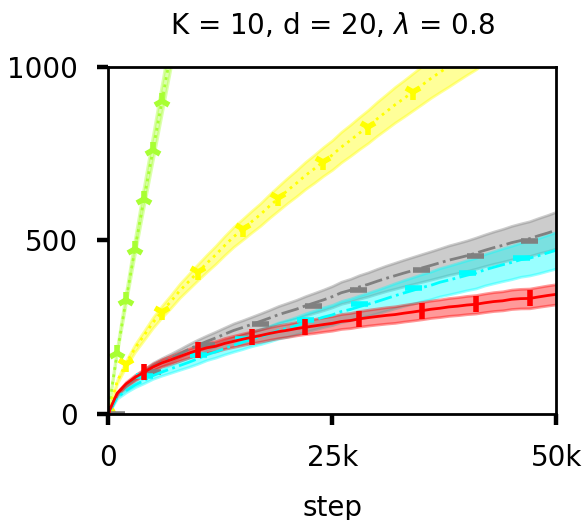}
	\includegraphics{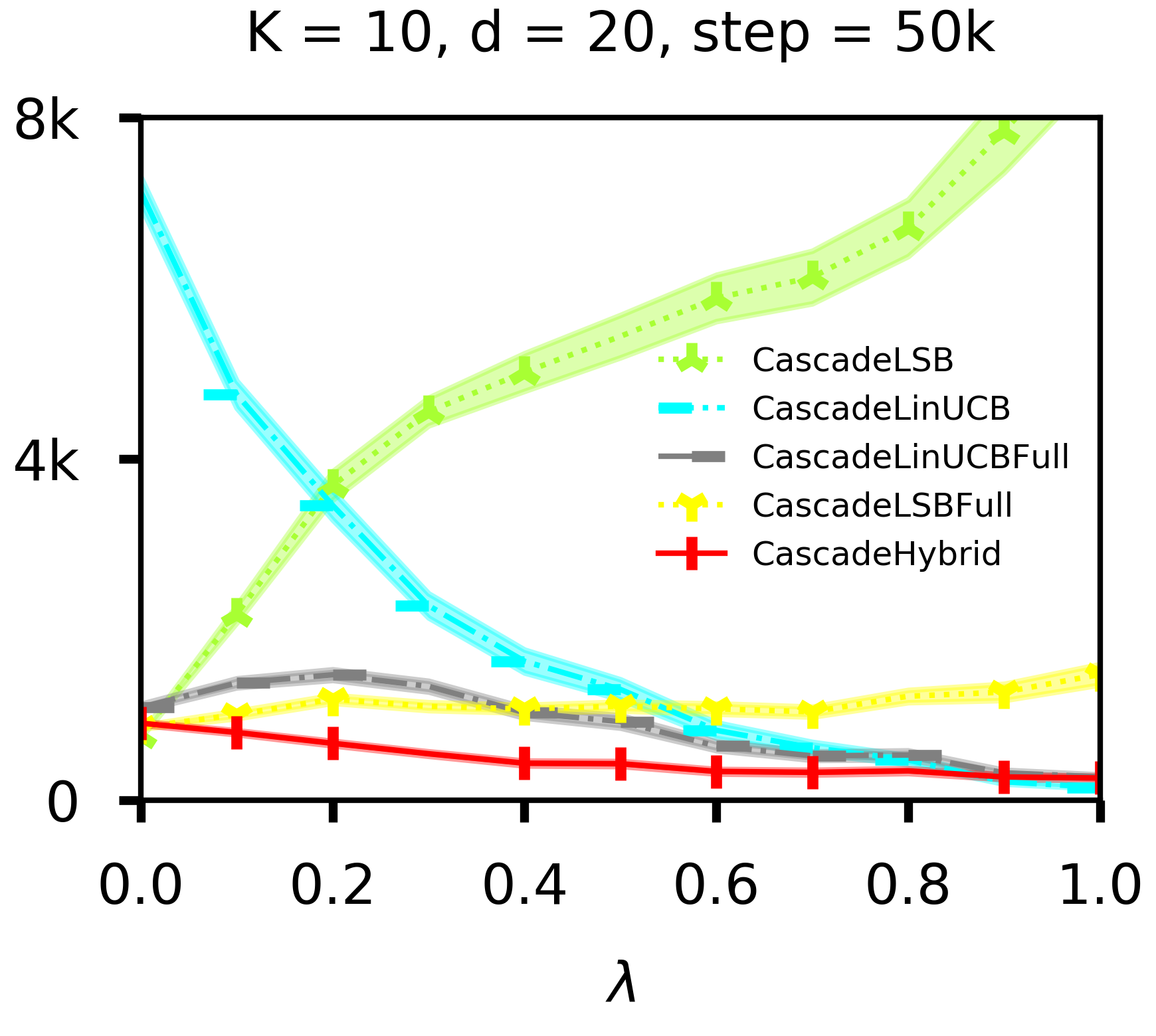}
	\includegraphics{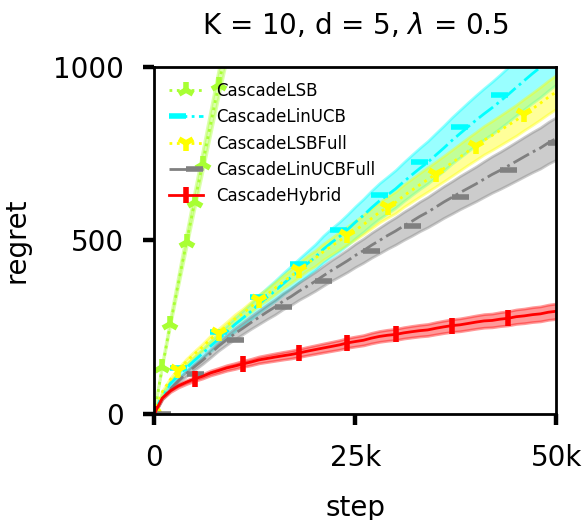}
	\includegraphics{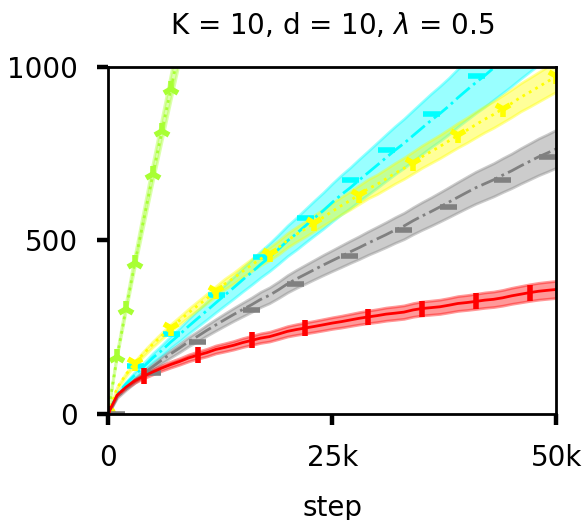}
	\includegraphics{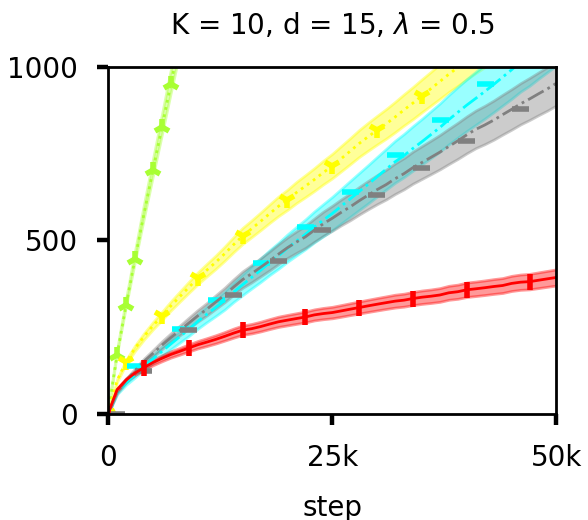}
	\includegraphics{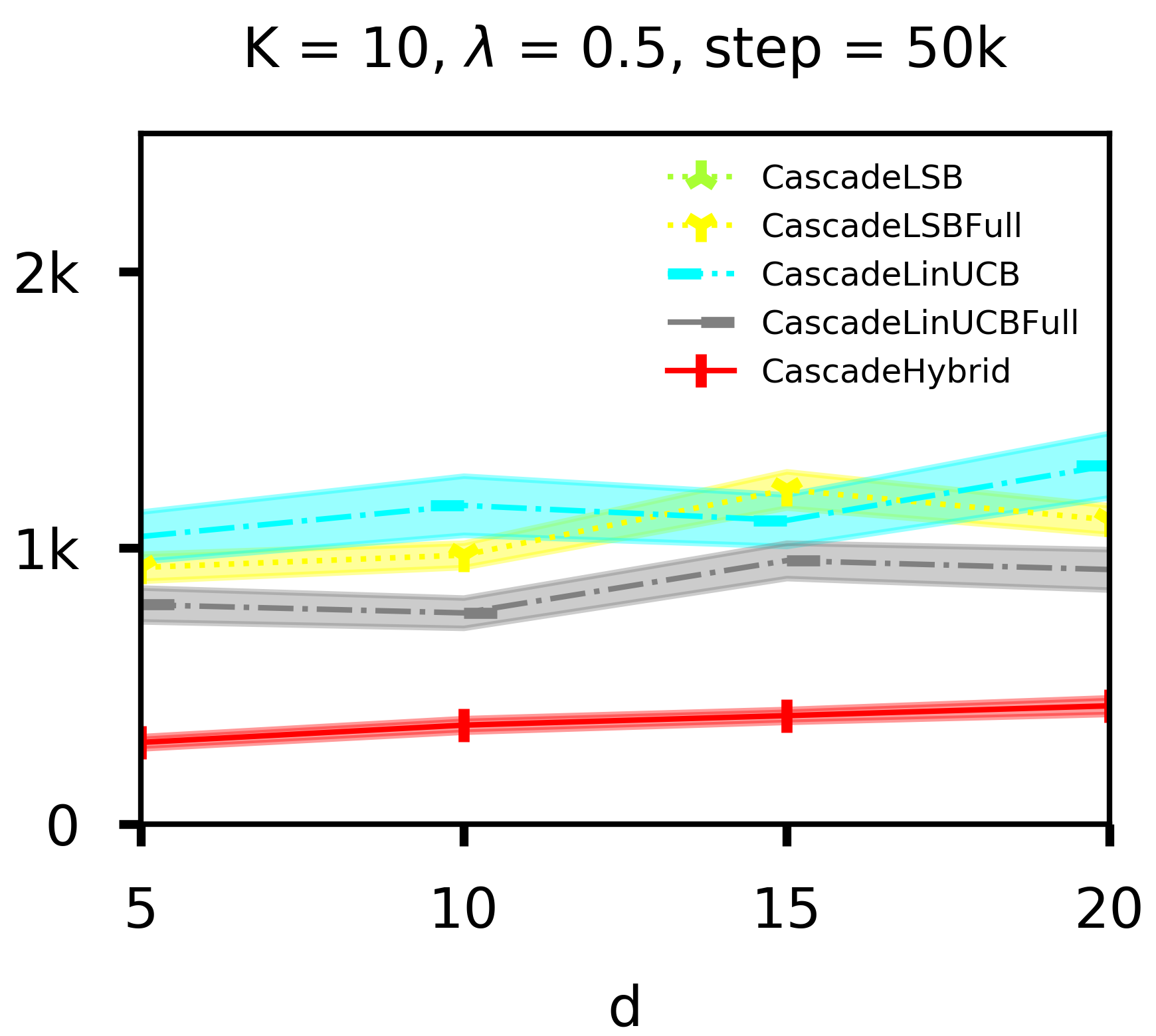}
	\includegraphics{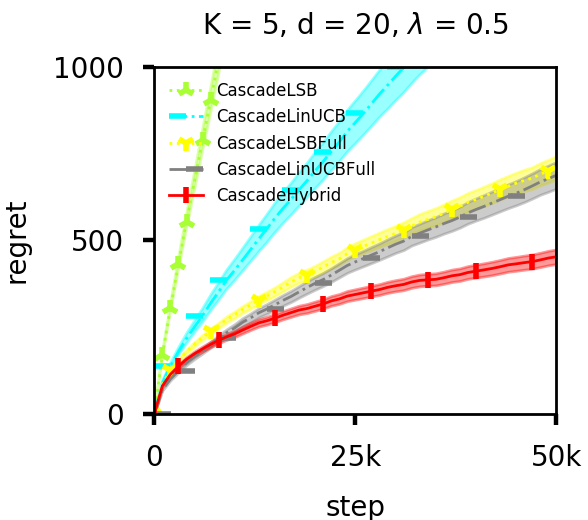}
	\includegraphics{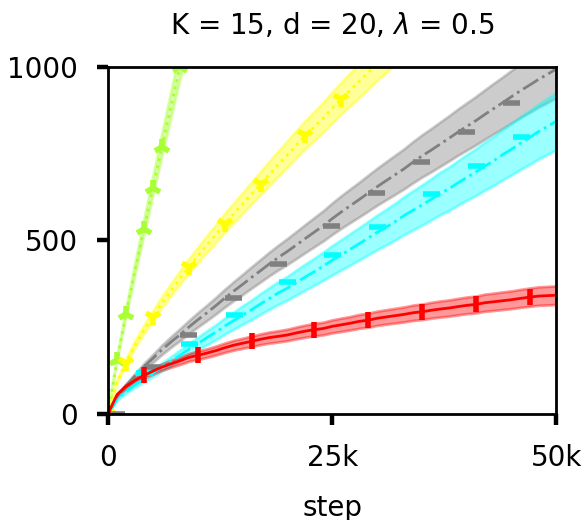}
	\includegraphics{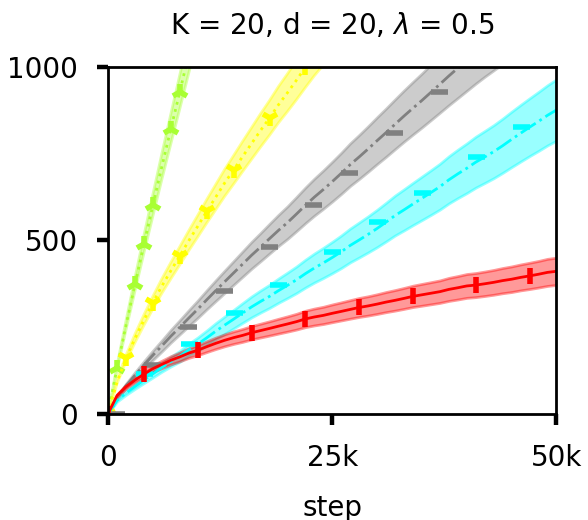}
	\includegraphics{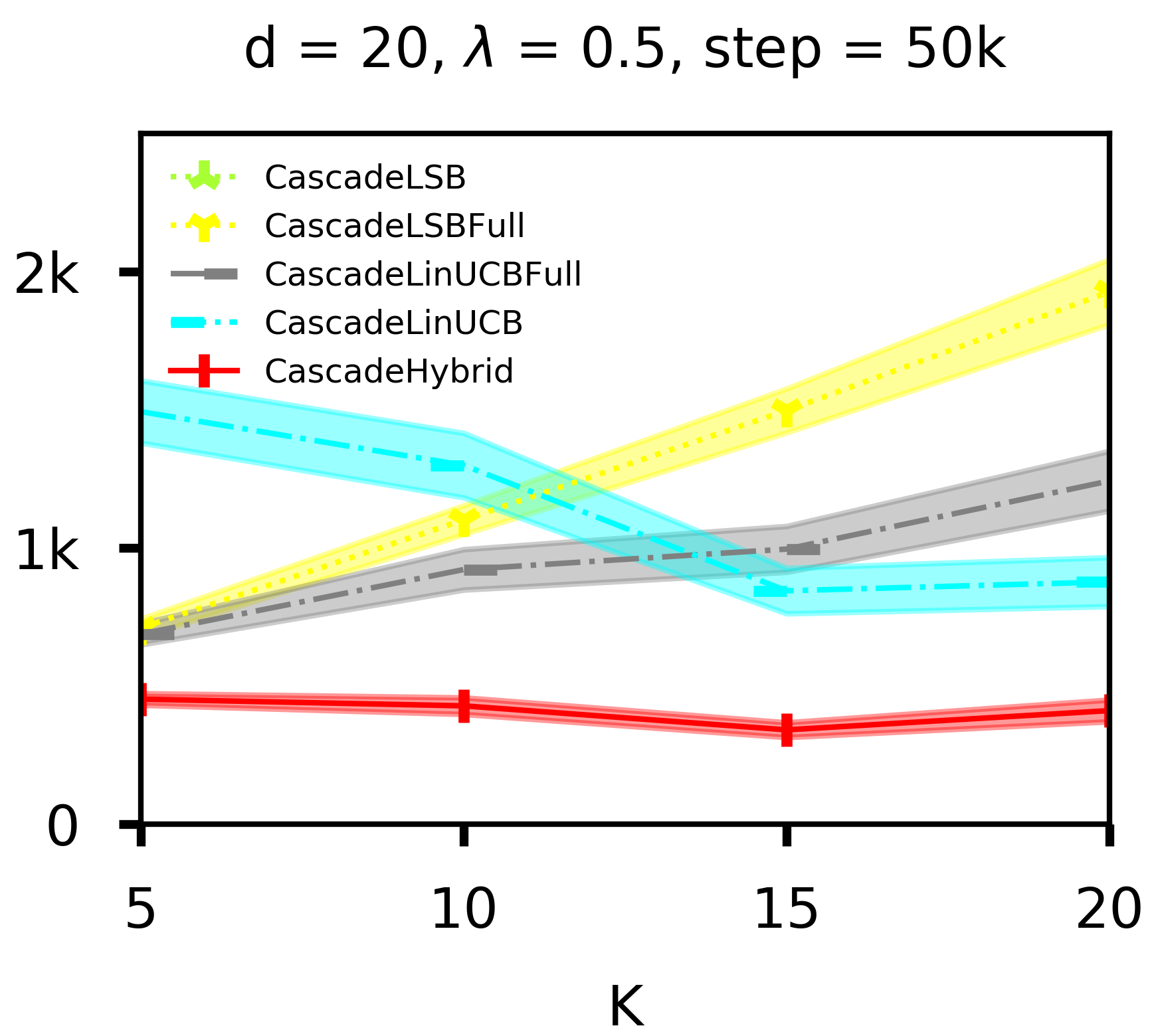}
	\caption{$n$-step regret on the Yahoo  dataset. Results are averaged over $500$ users with $2$ repeats per user. 
		Lower regret means more clicks received by the algorithm during the online learning. 
		Shaded areas are the standard errors of estimates. }  
	\label{fig:yahoo}	
\end{figure*}

We first study the movie recommendation task on the MovieLens dataset and show the results in \cref{fig:movie}.
The top row shows the impact of $\lambda$, where we fix $K=10$ and $d=20$. 
$\lambda$ is a trade-off parameter  in our simulation. 
It balances the relevance and diversity, and is unknown to online algorithms. 
Choosing a small $\lambda$, the simulated user prefers recommended movies in the list to be relevant, while choosing a large $\lambda$, the simulated user prefers the recommended movies to be diverse. 
As shown in the top row, $\cascadehybrid$ outperforms all baselines when $\lambda \in \{0.1, 0.2, \ldots 0.8\}$, and only loses to the particularly designed baselines ($\cascadelsb$ and $\cascadelinucb$) with small gaps in some extreme cases where they benefit most. 
This is reasonable since they have fewer parameters to be estimated than $\cascadehybrid$. 
In all cases, $\cascadehybrid$ has lower regret than $\cascadelinucbfull$ and $\cascadelsbfull$ that work with the same features as $\cascadehybrid$. 
This result indicates that including more features in $\cascadelsb$ and $\cascadelinucb$ is not sufficient to capture both item relevance and result diversification.  

The second row in \cref{fig:movie} shows the impact of different numbers of topics, where we fix $\lambda=0.5$ and $K=10$. 
We see that $\cascadehybrid$ outperforms all baselines with large gaps. 
In the last plot of the middle row, we see that the gap of regret between $\cascadehybrid$ and $\cascadelinucbfull$ decreases with larger values of $d$. 
This is because, on the MovieLens dataset, when $d$ is small, a user tends to prefer a diverse ranked list: when $d$ is small, an item is more likely to  belong to only one topic, and each entry of $\vtheta_j^*$ becomes relatively larger since $\sum_{j\in[d]}\vtheta_j^* = 1$. 
And given an item $a$ and a set $\cS$, the difference between $\Delta(a | \cS)^T \vtheta^*$ and $\Delta(a | \emptyset)^T \vtheta^*$ is large. 
This behavior is also confirmed by the fact that $\cascadelinucbfull$ outperforms $\cascadelsbfull$ for large $d$ while they perform similarly for small $d$. 
Finally, we study the impact of the number of positions on the regret. 
The results are displayed in the bottom row in \cref{fig:movie}, where we choose $\lambda=0.5$ and $d=20$. 
Again, we see that $\cascadehybrid$ outperforms baselines with large gaps. 

Next, we report the results on the Yahoo dataset in \cref{fig:yahoo}. 
We follow the same setup as for the MovieLens dataset and observe a similar behavior. 
$\cascadehybrid$ has slightly higher regret than the best performing baselines in three cases: $\cascadelsb$ when $\lambda=0$ and $\cascadelinucb$ when $\lambda\in\{0.9, 1\}$. 
Note that these are relatively extreme cases, where the particularly designed baselines can benefit most. 
Meanwhile, $\cascadelsb$ and $\cascadelinucb$ do not generalize well with different $\lambda$s. 
In all setups, $\cascadehybrid$ has lower regret than $\cascadelsbfull$ and $\cascadelinucbfull$, which confirms our hypothesis that the hybrid model has benefit in capturing both relevance and diversity.

%% file: sections/5-Analysis.tex

\section{Analysis}
\label{sec:analysis}

\subsection{Performance guarantee}
Since $\cascadehybrid$ competes with the greedy benchmark, we focus on the $\eta$-scaled expected $n$-step regret which is defined as: 
\begin{equation}
\label{eq:scaled regret}
R_\eta(n) = \sum_{t=1}^{n}\expect{  \eta r(\cR^*, \valpha) - r(\cR_t, A_t) }, 
\end{equation} 
where $\eta = (1-\frac{1}{e}) \max\{\frac{1}{K}, 1-\frac{K-1}{2}\alpha_{max}\}$.
This is a reasonable metric, since computing the optimal $\cR^*$ is computationally inefficient. 
A similar scaled regret has previously been used in diversity problems~\citep{yue-2011-linear,Qin2014Contextual,Hiranandani2019CascadingLS}. 
For simplicity, we write $\vw^* = [\vtheta^{*T}, \vbeta^{*T}]^T$.
Then, we bound the $\eta$-scaled regret of $\cascadehybrid$ as follows: 

\begin{thm}
\label{th: regret}
For  $\norm{\vw^*} \leq 1$ and any 
\begin{equation}
\label{eq:gamma}
\gamma \geq \sqrt{(m+d)\log\left(1+\frac{nK}{m+d}\right) + 2\log(n)} + \norm{\vw^*}, 
\end{equation}
we have 
\begin{equation}
	\label{eq:regret bound}
	R_{\eta}(n) \leq 
2\gamma \sqrt{{2nK(m+d)\log{\left(1+\frac{nK}{m+d}\right)}}}
	+ 1.	
\end{equation}
\end{thm}

\noindent%
Combining \cref{eq:gamma,eq:regret bound}, we have $R_\eta(n) = \tilde{O}((m+d)\sqrt{Kn})$, where the $\tilde{O}$ notation ignores logarithmic factors.   
Our bound has three characteristics: 
\begin{enumerate*}
	\item Theorem~\ref{th: regret} states a gap-free bound, where the factor $\sqrt{n}$ is considered near optimal;
	\item  This bound is linear in the number of features, which is a common dependence in learning bandit algorithms~\citep{abbasi2011improved}; and 
	\item Our bound is $\tilde{O}(\sqrt{K})$ lower than other bounds for linear bandit algorithms in \ac{CB}~\citep{zong16cascading,Hiranandani2019CascadingLS}. 
\end{enumerate*}
We include a proof of Theorem~\ref{th: regret} in \cref{sec:proof of theorem}. 
We use a similar strategy to decompose the regret as in~\citep{zong16cascading,Hiranandani2019CascadingLS}, but we have a better analysis on how to sum up the regret of individual items. 
Thus, our bound depends on  $\tilde{O}(\sqrt{K})$ rather than $\tilde{O}(K)$. 
We believe that our analysis can be applied to both $\cascadelinucb$ and $\cascadelsb$, and then show that their regret is actually bounded by $\tilde{O}(\sqrt{K})$ rather than $\tilde{O}(K)$.

\subsection{Proof of Theorem~\ref{th: regret}} 
\label{sec:proof of theorem}
We first define some additional notation. 
We write $\vw^* = [\vtheta^{*T}, \vbeta^{*T}]^T$. 
Given a ranked list $\cR$ and $a = \cR(i)$, we write $\vphi_a  = [\vomega_{a}^T, \vz_{a}^T]^T$.
With the $\vphi_a$ and $\vw^*$ notation, $\cascadehybrid$ can be viewed as an extension of $\cascadelsb$, where two submodular functions instead of one are used in a single model. 
We write $\vOt = \vI_{m+d} + \sum_{i=1}^{t-1}\sum_{a\in\cO_i} \vphi_a \vphi_a^T$ as the collected features in $t$ steps,  $\cH_t$ as the collected features and clicks up to step $t$, and $\cR^i = (\cR(1), \ldots, \cR(i))$.  
Then, the confidence bound in \cref{eq:cb} on the $i$-th item in $\cR$ can be re-written as:  
\begin{equation}
\label{eq:new cb}
s(\cR^i) = \vphi_{\cR(i)}^T \vOt^{-1} \vphi_{\cR(i)}. 
\end{equation}
Let $\Pi(\cD) = \bigcup_{i=1}^L\Pi_i(\cD)$ be the set of all ranked lists of $\cD$ with length $[L]$, and $\kappa: \Pi(\cD) \rightarrow [0, 1]$ be an arbitrary list function. 
For any $\cR \in \Pi(\cD)$ and any $\kappa$, we  define 
\begin{equation}
\label{eq:new f}
f(\cR, \kappa) = 1 - \prod_{i=1}^{|\cR|} (1-\kappa(\cR^i)).
\end{equation}
We define upper and lower confidence bounds, and $\kappa$ as: 
\begin{equation}
\label{eq:ucb lcb}
\begin{split}
u_t(\cR) ={}& \text{F}_{[0, 1]}[\vphi_{\cR(l)}^T\hat{\vw}_t + s(\cR^l)]  \\
l_t(\cR) = {}&\text{F}_{[0, 1]}[\vphi_{\cR(l)}^T\hat{\vw}_t - s(\cR^l)] \\
\kappa(R) = {}&\vphi_{\cR(l)}^T\vw^*, 
\end{split}
\end{equation}
where $l = |\cR|$ and  $\text{F}_{[0,1]}[\cdot] =  \max(0, \min(1, \cdot))$. 
With the definitions in \cref{eq:ucb lcb}, $f(\cR, \kappa) = r(\cR, \valpha)$ is the reward of list $\cR$.

\begin{proof}
	Let $g_t = \{ l_t(\cR) \leq \kappa(\cR) \leq u_t(\cR), \forall \cR \in \Pi(\cD)\}$ be the event that the attraction probabilities are bounded  by the lower and upper confidence bound, and $\bar{g}_t$ be the complement of $g_t$. 
	We have 
	\begin{eqnarray}
	\label{eq:regret to f}
	\begin{split}
	&\expect{\eta r(\cR^*, \valpha) - r(\cR_t, \vA_t)}  =	\expect{\eta f(\cR^*, \kappa) - f(\cR_t, \kappa)} \\
	&
	\stackrel{(a)}{\leq} P(g_t) \expect{\eta f(\cR^*, \kappa) - f(\cR_t, \kappa)} + P(\bar{g}_t) \\
	&
	\stackrel{(b)}{\leq} P(g_t) \expect{\eta f(\cR^*, u_t) - f(\cR_t, \kappa)} + P(\bar{g}_t)  
	\\
	&
	\stackrel{(c)}{\leq} P(g_t) \expect{ f(\cR_t, u_t) - f(\cR_t, \kappa)} + P(\bar{g}_t),  
	\end{split}
	\end{eqnarray}
	where $(a)$ holds because $\expect{\eta f(\cR^*, \kappa) - f(\cR_t, \kappa)} \leq 1$,  $(b)$ holds because under event $g_t$ we have $f(\cR, l_t) \leq f(\cR, \kappa) \leq f(\cR, u_t)$, $\forall \cR \in \Pi(\cD)$, and $(c)$ holds by the definition of the $\eta$-approximation, where we have  
	\begin{equation}
	\eta f(\cR^*, u_t) \leq  \max_{\cR \in \Pi_K(\cD)} \eta f(\cR, u_t) \leq f(\cR_t, u_t). 
	\end{equation}
	By the definition of the list function $f(\cdot, \cdot)$ in \cref{eq:new f}, we have 
		\begin{equation}
		\begin{split}
		&f(\cR_t, u_t) - f(\cR_t, \kappa)  \\
		&= \prod_{k=1}^{K} (1- \kappa(\cR_t^k)) -  \prod_{k=1}^{K} (1- u_t(\cR_t^k)) 
		\\
		&\stackrel{(a)}{=} \sum_{k=1}^{K} \!\!\left[ \prod_{i=1}^{k-1} (1- \kappa(\cR_t^i)) \right] \!\!(u_t(\cR_t^k) - \kappa(\cR_t^k))\!\!
		\left[ \prod_{j=k+1}^{K} (1- u(\cR_t^j)) \right]   
		\\&
		\stackrel{(b)}{\leq}  \sum_{k=1}^{K}\!\! \left[ \prod_{i=1}^{k-1} (1- \kappa(\cR_t^i)) \right] \!\!(u_t(\cR_t^k) - \kappa(\cR_t^k)),
		\end{split}
		\end{equation}%
	where $(a)$ follows from Lemma 1 in~\citep{zong16cascading} and $(b)$ is because of the fact that $0\leq \kappa(\cR_t) \leq u_t(\cR_t) \leq 1$. 
	We then define the event $h_{ti} = \{\text{item } \cR_t(i) \text{ is observed} \}$, where we have $\expect{\ind{h_{ti}}} = \prod_{k=1}^{i-1} (1- \kappa(\cR_t^k))$.
	For any $\cH_t$ such that $g_t$ holds, we have 
	\begin{equation}
	\label{eq:f to sqrt}
	\begin{split}
	&  \expect{f(\cR_t, u_t) - f(\cR_t, \kappa) \mid \cH_t} \\  
	& \leq \sum_{i=1}^K \expect{\ind{h_{ti}} \mid \cH_t }(u_t(\cR_t^i) - l_t(\cR_t^i))  \\
	&\stackrel{(a)}{\leq} 2\gamma \expect{\ind{h_{ti}} \sum_{i=1}^{K}\sqrt{s(\cR_t^i)}   \mid \cH_t}  
	\stackrel{(b)}{\leq} 2\gamma \expect{\sum_{i=1}^{\min(K, c_t)}\sqrt{s(\cR_t^i)}   \mid \cH_t},  
	\end{split}	
	\end{equation}
	where inequality $(a)$ follows from the definition of $u_t$ and $l_t$ in \cref{eq:ucb lcb}, and inequality $(b)$ follows from the definition of $h_{ti}$.
	Now, together with \cref{eq:scaled regret,eq:regret to f,eq:f to sqrt}, we have 
	\begin{equation}
	\label{eq:proof cumulative regret}
	\begin{split}
	R_{\eta}(n)& = \sum_{t=1}^{n}\expect{\eta r(\cR^*, \valpha) - r(\cR_t, \vA_t)  }  \\
	&\leq  \sum_{t=1}^{n} \left[ 2\gamma  \expect{\sum_{i=1}^{\min(K, c_t)}\sqrt{s(\cR_t^i)}   \mid g_t} P(g_t) 
	+ P(\bar{g}_t) 
	\right] 
	\\ &
	\leq     2\gamma  \expect{\sum_{t=1}^{n}\sum_{i=1}^{K}\sqrt{s(\cR_t^i)} }
	+ \sum_{t=1}^{n}p(\bar{g}_t).  
	\end{split}
	\end{equation}
	For the first term in \cref{eq:proof cumulative regret}, we have
	\begin{equation}
	\begin{split}
	\label{eq:to det}
	&\sum_{t=1}^{n}\sum_{i=1}^{K}\sqrt{s(\cR_t^i)}
	\stackrel{(a)}{\leq} \sqrt{nK \sum_{t=1}^{n}\sum_{i=1}^{K}s(\cR_t^i)} 
	\stackrel{(b)}{\leq} 
	\sqrt{nK 2\log det(\vO_t) },  
	\end{split}	
	\end{equation}
	where inequality $(a)$ follows from the Cauchy-Schwarz inequality and $(b)$ follows from Lemma 5 in~\citep{yue-2011-linear}. 
	Note that $\log det(\vO_t) \leq (m+d) \log(K(1+n/(m+d)))$, which can be obtained by the determinant and trace inequality, and together with \cref{eq:to det}:
	\begin{equation}
	\label{eq:to log}
	\mbox{}\hspace*{-2mm}
	\sum_{t=1}^{n}\sum_{i=1}^{K}\sqrt{s(\cR_t^i)} 
	\leq 
	\sqrt{2nK(m+d)\log(K(1+\frac{n}{m+d}))  }. 
	\hspace*{-2mm}\mbox{}
	\end{equation}
	For the second term in \cref{eq:proof cumulative regret}, by Lemma 3 in~\citep{Hiranandani2019CascadingLS}, we have $P(\bar{g}_t) \leq 1/n$ for any $\gamma$ that satisfies \cref{eq:gamma}. 
	Thus, together with \cref{eq:proof cumulative regret,eq:to det,eq:to log}, we have 
	\begin{equation*}
	R_{\eta}(n) \leq 2\gamma\sqrt{2nK(m+d)\log(1+\frac{nK}{m+d}) }  +1. 
	\end{equation*}
	This concludes the proof of Theorem~\ref{th: regret}.
\end{proof}

%% file: sections/6-RelatedWork.tex

\section{Related Work}
\label{sec:related work}

The literature on offline \acf{LTR} methods that account for position bias and diversity is too broad to review in detail. 
We refer readers to \cite{aggarwal2016recommender} for an overview. 
In this section, we mainly review online \ac{LTR} papers that are closely related to our work, i.e., stochastic click bandit models.  

Online \ac{LTR} in a stochastic click models has been well-studied~\citep{radlinski08learning,slivkins2010learning,yue-2011-linear, kveton15cascading,katariya16dcm,zong16cascading,batchrank,li-2019-online,li-2019-cascading,li2019bubblerank}. 
Previous work can be categorized into two groups: feature-free models and feature-rich models. 
Algorithms from the former group use a tabular representation on items and maintain an estimator for each item. 
They learn inefficiently and are limited to the problem with a small number of item candidates. 
In this paper, we focus on the ranking problem with a large number of items. 
Thus, we do not consider feature-free model in the experiments. 

Feature-rich models learn efficiently in terms of the number of items.
They are suitable for large-scale ranking problems. 
Among them, ranked bandits~\citep{radlinski08learning,slivkins2010learning} are early approaches to online \ac{LTR}. 
In ranked bandits, each position is model as a \ac{MAB} and diversity of results is addressed in the sense that items ranked at lower positions are less likely to be clicked than those at higher positions, which is different from the topical diversity as we study. 
Also, ranked bandits do not consider the position bias and are suboptimal in the problem where a user browse different possition unevenly, e.g., \ac{CM}~\citep{kveton15cascading}.
$\lsbgreedy$~\citep{yue-2011-linear} and $\ccucb$~\citep{Qin2014Contextual} use submodular functions to solve the online diverse \ac{LTR} problem. 
They assume that the user browses all displayed items and, thus, do not consider the position bias either.

Our work is closely related to $\cascadelinucb$~\citep{zong16cascading} and  \break  $\cascadelsb$~\citep{Hiranandani2019CascadingLS},  the baselines in our experiments, and can be viewed as a combination of both. 
$\cascadelinucb$ solves the relevance ranking in the \ac{CM} and assumes the attraction probability is a linear combination of features. 
$\cascadelsb$ is designed for result diversification and assumes that the attraction probability is computed as a submodular function; see \cref{eq:topic coverage attration probability}. 
In our $\cascadehybrid$, the attraction probability is a hybrid of both; see \cref{eq:hybridy attraction probability}. 
Thus, $\cascadehybrid$ handles both relevance ranking and result diversification. 
$\recurrank$~\citep{li-2019-online} is a recently proposed algorithm that aims at learning the optimal list in term of item relevance in most click models. 
However, to achieve this task, $\recurrank$ requires a lot of randomly shuffled lists and is outperformed by $\cascadelinucb$ in the \ac{CM}~\citep{li-2019-online}. 

The hybrid of a linear function and a submodular function has been used in solving combinatorial semi-bandits.
\citet{perrault19a-exploiting} use a linear set function to model the expected reward of arm, and use the submodular function to compute the \emph{exploration bonus}. 
This is different from our hybrid model, where both the linear and submodular functions are used to model the attraction probability and the confident bound is used as the exploration bonus. 

%% file: sections/7-Conclusions.tex

\section{Conclusion}
\label{sec:conclusion}

In real world interactive systems, both relevance of individual items and topical diversity of result lists are critical factors in user satisfaction. 
In order to better meet users' information needs, we propose a novel online \ac{LTR} algorithm that optimizes both factors in a hybrid fashion.  
We formulate the problem as \acf{CHB}, where the attraction probability is a hybrid function that combines a function of relevance features and a submodular function of topic features. 
$\cascadehybrid$ utilizes a hybrid model as a scoring function and the \ac{UCB} policy for exploration. 
We provide a gap-free bound on the $\eta$-scaled $n$-step regret of $\cascadehybrid$, and conduct experiments on two real-world datasets. 
Our empirical study shows that $\cascadehybrid$ outperforms two existing online \ac{LTR} algorithms that exclusively consider either relevance ranking or result diversification.

In future work, we intend to conduct experiments on live systems, where feedback is obtained from multiple users so as to test whether $\cascadehybrid$ can learn across users. 
Another direction is to adapt Thompson sampling~\citep{thompson-1933} to our hybrid model, since Thompson sampling generally outperforms \ac{UCB}-based algorithms~\citep{zong16cascading,li-2020-mergedts}. 

%% file: paper.bbl

\begin{thebibliography}{35}


\ifx \showCODEN    \undefined \def \showCODEN     #1{\unskip}     \fi
\ifx \showDOI      \undefined \def \showDOI       #1{#1}\fi
\ifx \showISBNx    \undefined \def \showISBNx     #1{\unskip}     \fi
\ifx \showISBNxiii \undefined \def \showISBNxiii  #1{\unskip}     \fi
\ifx \showISSN     \undefined \def \showISSN      #1{\unskip}     \fi
\ifx \showLCCN     \undefined \def \showLCCN      #1{\unskip}     \fi
\ifx \shownote     \undefined \def \shownote      #1{#1}          \fi
\ifx \showarticletitle \undefined \def \showarticletitle #1{#1}   \fi
\ifx \showURL      \undefined \def \showURL       {\relax}        \fi
\providecommand\bibfield[2]{#2}
\providecommand\bibinfo[2]{#2}
\providecommand\natexlab[1]{#1}
\providecommand\showeprint[2][]{arXiv:#2}

\bibitem[\protect\citeauthoryear{Abbasi-Yadkori, P{\'a}l, and
  Szepesv{\'a}ri}{Abbasi-Yadkori et~al\mbox{.}}{2011}]%
        {abbasi2011improved}
\bibfield{author}{\bibinfo{person}{Yasin Abbasi-Yadkori},
  \bibinfo{person}{D{\'a}vid P{\'a}l}, {and} \bibinfo{person}{Csaba
  Szepesv{\'a}ri}.} \bibinfo{year}{2011}\natexlab{}.
\newblock \showarticletitle{Improved Algorithms for Linear Stochastic Bandits}.
  In \bibinfo{booktitle}{\emph{NIPS}}. \bibinfo{pages}{2312--2320}.
\newblock


\bibitem[\protect\citeauthoryear{Aggarwal}{Aggarwal}{2016}]%
        {aggarwal2016recommender}
\bibfield{author}{\bibinfo{person}{Charu~C. Aggarwal}.}
  \bibinfo{year}{2016}\natexlab{}.
\newblock \bibinfo{booktitle}{\emph{Recommender Systems: The Textbook}}.
\newblock \bibinfo{publisher}{Springer}.
\newblock


\bibitem[\protect\citeauthoryear{Agrawal, Gollapudi, Halverson, and
  Ieong}{Agrawal et~al\mbox{.}}{2009}]%
        {Agrawal2009-diversifying}
\bibfield{author}{\bibinfo{person}{Rakesh Agrawal}, \bibinfo{person}{Sreenivas
  Gollapudi}, \bibinfo{person}{Alan Halverson}, {and} \bibinfo{person}{Samuel
  Ieong}.} \bibinfo{year}{2009}\natexlab{}.
\newblock \showarticletitle{Diversifying Search Results}. In
  \bibinfo{booktitle}{\emph{WSDM}}. \bibinfo{pages}{5--14}.
\newblock


\bibitem[\protect\citeauthoryear{Ashkan, Kveton, Berkovsky, and Wen}{Ashkan
  et~al\mbox{.}}{2015}]%
        {Ashkan2015-optimal}
\bibfield{author}{\bibinfo{person}{Azin Ashkan}, \bibinfo{person}{Branislav
  Kveton}, \bibinfo{person}{Shlomo Berkovsky}, {and} \bibinfo{person}{Zheng
  Wen}.} \bibinfo{year}{2015}\natexlab{}.
\newblock \showarticletitle{Optimal Greedy Diversity for Recommendation}. In
  \bibinfo{booktitle}{\emph{IJCAI}}. \bibinfo{pages}{1742--1748}.
\newblock


\bibitem[\protect\citeauthoryear{Auer, Cesa-Bianchi, and Fischer}{Auer
  et~al\mbox{.}}{2002}]%
        {auer02finitetime}
\bibfield{author}{\bibinfo{person}{Peter Auer}, \bibinfo{person}{Nicolo
  Cesa-Bianchi}, {and} \bibinfo{person}{Paul Fischer}.}
  \bibinfo{year}{2002}\natexlab{}.
\newblock \showarticletitle{Finite-time Analysis of the Multiarmed Bandit
  Problem}.
\newblock \bibinfo{journal}{\emph{Machine Learning}}  \bibinfo{volume}{47}
  (\bibinfo{year}{2002}), \bibinfo{pages}{235--256}.
\newblock


\bibitem[\protect\citeauthoryear{Chuklin, Markov, and de~Rijke}{Chuklin
  et~al\mbox{.}}{2015}]%
        {chuklin2015click}
\bibfield{author}{\bibinfo{person}{Aleksandr Chuklin}, \bibinfo{person}{Ilya
  Markov}, {and} \bibinfo{person}{Maarten de Rijke}.}
  \bibinfo{year}{2015}\natexlab{}.
\newblock \bibinfo{booktitle}{\emph{Click Models for Web Search}}.
\newblock \bibinfo{publisher}{Morgan \& Claypool}.
\newblock


\bibitem[\protect\citeauthoryear{Craswell, Zoeter, Taylor, and Ramsey}{Craswell
  et~al\mbox{.}}{2008}]%
        {craswell08experimental}
\bibfield{author}{\bibinfo{person}{Nick Craswell}, \bibinfo{person}{Onno
  Zoeter}, \bibinfo{person}{Michael Taylor}, {and} \bibinfo{person}{Bill
  Ramsey}.} \bibinfo{year}{2008}\natexlab{}.
\newblock \showarticletitle{An Experimental Comparison of Click Position-Bias
  Models}. In \bibinfo{booktitle}{\emph{WSDM}}. \bibinfo{pages}{87--94}.
\newblock


\bibitem[\protect\citeauthoryear{Golub and Van~Loan}{Golub and
  Van~Loan}{1996}]%
        {golub-1996-matrix}
\bibfield{author}{\bibinfo{person}{Gene~H. Golub} {and}
  \bibinfo{person}{Charles~F. Van~Loan}.} \bibinfo{year}{1996}\natexlab{}.
\newblock \bibinfo{booktitle}{\emph{Matrix Computations} (\bibinfo{edition}{3}
  ed.)}.
\newblock \bibinfo{publisher}{Johns Hopkins}, \bibinfo{address}{Baltimore, MD}.
\newblock


\bibitem[\protect\citeauthoryear{Grotov and de~Rijke}{Grotov and
  de~Rijke}{2016}]%
        {grotov-2016-online}
\bibfield{author}{\bibinfo{person}{Artem Grotov} {and} \bibinfo{person}{Maarten
  de Rijke}.} \bibinfo{year}{2016}\natexlab{}.
\newblock \showarticletitle{Online Learning to Rank for Information Retrieval:
  SIGIR 2016 Tutorial}. In \bibinfo{booktitle}{\emph{SIGIR}}.
  \bibinfo{pages}{1215--1218}.
\newblock


\bibitem[\protect\citeauthoryear{Harper and Konstan}{Harper and
  Konstan}{2015}]%
        {Harper:2015:movielense}
\bibfield{author}{\bibinfo{person}{F.~Maxwell Harper} {and}
  \bibinfo{person}{Joseph~A. Konstan}.} \bibinfo{year}{2015}\natexlab{}.
\newblock \showarticletitle{The MovieLens Datasets: History and Context}.
\newblock \bibinfo{journal}{\emph{ACM Trans. Interact. Intell. Syst.}}
  \bibinfo{volume}{5}, \bibinfo{number}{4} (\bibinfo{year}{2015}),
  \bibinfo{pages}{19:1--19:19}.
\newblock


\bibitem[\protect\citeauthoryear{Hiranandani, Singh, Gupta, Burhanuddin, Wen,
  and Kveton}{Hiranandani et~al\mbox{.}}{2019}]%
        {Hiranandani2019CascadingLS}
\bibfield{author}{\bibinfo{person}{Gaurush Hiranandani},
  \bibinfo{person}{Harvineet Singh}, \bibinfo{person}{Prakhar Gupta},
  \bibinfo{person}{Iftikhar~Ahamath Burhanuddin}, \bibinfo{person}{Zheng Wen},
  {and} \bibinfo{person}{Branislav Kveton}.} \bibinfo{year}{2019}\natexlab{}.
\newblock \showarticletitle{Cascading Linear Submodular Bandits: Accounting for
  Position Bias and Diversity in Online Learning to Rank}. In
  \bibinfo{booktitle}{\emph{UAI}}.
\newblock


\bibitem[\protect\citeauthoryear{Hofmann, Schuth, Whiteson, and
  de~Rijke}{Hofmann et~al\mbox{.}}{2013}]%
        {hofmann2013reusing}
\bibfield{author}{\bibinfo{person}{Katja Hofmann}, \bibinfo{person}{Anne
  Schuth}, \bibinfo{person}{Shimon Whiteson}, {and} \bibinfo{person}{Maarten de
  Rijke}.} \bibinfo{year}{2013}\natexlab{}.
\newblock \showarticletitle{Reusing Historical Interaction Data for Faster
  Online Learning to Rank for {IR}}. In \bibinfo{booktitle}{\emph{WSDM}}.
  \bibinfo{pages}{183--192}.
\newblock


\bibitem[\protect\citeauthoryear{Hofmann, Whiteson, and de~Rijke}{Hofmann
  et~al\mbox{.}}{2011}]%
        {hofmann-balancing-2011}
\bibfield{author}{\bibinfo{person}{Katja Hofmann}, \bibinfo{person}{Shimon
  Whiteson}, {and} \bibinfo{person}{Maarten de Rijke}.}
  \bibinfo{year}{2011}\natexlab{}.
\newblock \showarticletitle{Balancing exploration and exploitation in learning
  to rank online}. In \bibinfo{booktitle}{\emph{ECIR}}.
  \bibinfo{pages}{251--263}.
\newblock


\bibitem[\protect\citeauthoryear{Jagerman, Oosterhuis, and de~Rijke}{Jagerman
  et~al\mbox{.}}{2019}]%
        {jagerman2019comparison}
\bibfield{author}{\bibinfo{person}{Rolf Jagerman}, \bibinfo{person}{Harrie
  Oosterhuis}, {and} \bibinfo{person}{Maarten de Rijke}.}
  \bibinfo{year}{2019}\natexlab{}.
\newblock \showarticletitle{To Model or to Intervene: A Comparison of
  Counterfactual and Online Learning to Rank from User Interactions}. In
  \bibinfo{booktitle}{\emph{SIGIR}}. \bibinfo{pages}{15--24}.
\newblock


\bibitem[\protect\citeauthoryear{Katariya, Kveton, Szepesvari, and
  Wen}{Katariya et~al\mbox{.}}{2016}]%
        {katariya16dcm}
\bibfield{author}{\bibinfo{person}{Sumeet Katariya}, \bibinfo{person}{Branislav
  Kveton}, \bibinfo{person}{Csaba Szepesvari}, {and} \bibinfo{person}{Zheng
  Wen}.} \bibinfo{year}{2016}\natexlab{}.
\newblock \showarticletitle{{DCM} Bandits: Learning to Rank with Multiple
  Clicks}. In \bibinfo{booktitle}{\emph{ICML}}. \bibinfo{pages}{1215--1224}.
\newblock


\bibitem[\protect\citeauthoryear{Kveton, Szepesvari, Wen, and Ashkan}{Kveton
  et~al\mbox{.}}{2015}]%
        {kveton15cascading}
\bibfield{author}{\bibinfo{person}{Branislav Kveton}, \bibinfo{person}{Csaba
  Szepesvari}, \bibinfo{person}{Zheng Wen}, {and} \bibinfo{person}{Azin
  Ashkan}.} \bibinfo{year}{2015}\natexlab{}.
\newblock \showarticletitle{Cascading Bandits: Learning to Rank in the Cascade
  Model}. In \bibinfo{booktitle}{\emph{ICML}}. \bibinfo{pages}{767--776}.
\newblock


\bibitem[\protect\citeauthoryear{Lattimore and Szepesv{\'a}ri}{Lattimore and
  Szepesv{\'a}ri}{2020}]%
        {lattimore2018bandit}
\bibfield{author}{\bibinfo{person}{Tor Lattimore} {and} \bibinfo{person}{Csaba
  Szepesv{\'a}ri}.} \bibinfo{year}{2020}\natexlab{}.
\newblock \bibinfo{booktitle}{\emph{Bandit Algorithms}}.
\newblock \bibinfo{publisher}{Cambridge University Press}.
\newblock


\bibitem[\protect\citeauthoryear{Li and de~Rijke}{Li and de~Rijke}{2019}]%
        {li-2019-cascading}
\bibfield{author}{\bibinfo{person}{Chang Li} {and} \bibinfo{person}{Maarten de
  Rijke}.} \bibinfo{year}{2019}\natexlab{}.
\newblock \showarticletitle{Cascading Non-stationary Bandits: Online Learning
  to Rank in the Non-stationary Cascade Model}. In
  \bibinfo{booktitle}{\emph{IJCAI}}. \bibinfo{pages}{2859--2865}.
\newblock


\bibitem[\protect\citeauthoryear{Li, Kveton, Lattimore, Markov, de~Rijke,
  Szepesv{\'a}ri, and Zoghi}{Li et~al\mbox{.}}{2019a}]%
        {li2019bubblerank}
\bibfield{author}{\bibinfo{person}{Chang Li}, \bibinfo{person}{Branislav
  Kveton}, \bibinfo{person}{Tor Lattimore}, \bibinfo{person}{Ilya Markov},
  \bibinfo{person}{Maarten de Rijke}, \bibinfo{person}{Csaba Szepesv{\'a}ri},
  {and} \bibinfo{person}{Masrour Zoghi}.} \bibinfo{year}{2019}\natexlab{a}.
\newblock \showarticletitle{{BubbleRank}: Safe Online Learning to Re-Rank via
  Implicit Click Feedback}. In \bibinfo{booktitle}{\emph{UAI}}.
\newblock


\bibitem[\protect\citeauthoryear{Li, Markov, de~Rijke, and Zoghi}{Li
  et~al\mbox{.}}{2020}]%
        {li-2020-mergedts}
\bibfield{author}{\bibinfo{person}{Chang Li}, \bibinfo{person}{Ilya Markov},
  \bibinfo{person}{Maarten de Rijke}, {and} \bibinfo{person}{Masrour Zoghi}.}
  \bibinfo{year}{2020}\natexlab{}.
\newblock \showarticletitle{MergeDTS: A Method for Effective Large-scale Online
  Ranker Evaluation}.
\newblock \bibinfo{journal}{\emph{ACM Transactions on Information Systems}}
  \bibinfo{volume}{38}, \bibinfo{number}{4} (\bibinfo{date}{August}
  \bibinfo{year}{2020}).
\newblock


\bibitem[\protect\citeauthoryear{Li, Chu, Langford, and Schapire}{Li
  et~al\mbox{.}}{2010}]%
        {li2010contextual}
\bibfield{author}{\bibinfo{person}{Lihong Li}, \bibinfo{person}{Wei Chu},
  \bibinfo{person}{John Langford}, {and} \bibinfo{person}{Robert~E Schapire}.}
  \bibinfo{year}{2010}\natexlab{}.
\newblock \showarticletitle{A Contextual-bandit Approach to Personalized News
  Article Recommendation}. In \bibinfo{booktitle}{\emph{WWW}}.
  \bibinfo{pages}{661--670}.
\newblock


\bibitem[\protect\citeauthoryear{Li, Lattimore, and Szepesv{\'a}ri}{Li
  et~al\mbox{.}}{2019b}]%
        {li-2019-online}
\bibfield{author}{\bibinfo{person}{Shuai Li}, \bibinfo{person}{Tor Lattimore},
  {and} \bibinfo{person}{Csaba Szepesv{\'a}ri}.}
  \bibinfo{year}{2019}\natexlab{b}.
\newblock \showarticletitle{Online Learning to Rank with Features}. In
  \bibinfo{booktitle}{\emph{ICML}}. \bibinfo{pages}{3856--3865}.
\newblock


\bibitem[\protect\citeauthoryear{Liu}{Liu}{2009}]%
        {liu2009learning}
\bibfield{author}{\bibinfo{person}{Tie-Yan Liu}.}
  \bibinfo{year}{2009}\natexlab{}.
\newblock \showarticletitle{Learning to Rank for Information Retrieval}.
\newblock \bibinfo{journal}{\emph{Foundations and Trends in Information
  Retrieval}} \bibinfo{volume}{3}, \bibinfo{number}{3} (\bibinfo{year}{2009}),
  \bibinfo{pages}{225--331}.
\newblock


\bibitem[\protect\citeauthoryear{Nemhauser, Wolsey, and Fisher}{Nemhauser
  et~al\mbox{.}}{1978}]%
        {nemhauser1978analysis}
\bibfield{author}{\bibinfo{person}{George~L Nemhauser},
  \bibinfo{person}{Laurence~A Wolsey}, {and} \bibinfo{person}{Marshall~L
  Fisher}.} \bibinfo{year}{1978}\natexlab{}.
\newblock \showarticletitle{An Analysis of Approximations for Maximizing
  Submodular Set Functions---I}.
\newblock \bibinfo{journal}{\emph{Mathematical Programming}}
  \bibinfo{volume}{14}, \bibinfo{number}{1} (\bibinfo{year}{1978}),
  \bibinfo{pages}{265--294}.
\newblock


\bibitem[\protect\citeauthoryear{Oosterhuis and de~Rijke}{Oosterhuis and
  de~Rijke}{2018}]%
        {oosterhuis-differentiable-2018}
\bibfield{author}{\bibinfo{person}{Harrie Oosterhuis} {and}
  \bibinfo{person}{Maarten de Rijke}.} \bibinfo{year}{2018}\natexlab{}.
\newblock \showarticletitle{Differentiable Unbiased Online Learning to Rank}.
  In \bibinfo{booktitle}{\emph{CIKM}}. \bibinfo{pages}{1293--1302}.
\newblock


\bibitem[\protect\citeauthoryear{Perrault, Perchet, and Valko}{Perrault
  et~al\mbox{.}}{2019}]%
        {perrault19a-exploiting}
\bibfield{author}{\bibinfo{person}{Pierre Perrault}, \bibinfo{person}{Vianney
  Perchet}, {and} \bibinfo{person}{Michal Valko}.}
  \bibinfo{year}{2019}\natexlab{}.
\newblock \showarticletitle{Exploiting Structure of Uncertainty for Efficient
  Matroid Semi-bandits}. In \bibinfo{booktitle}{\emph{ICML}}.
  \bibinfo{publisher}{PMLR}, \bibinfo{pages}{5123--5132}.
\newblock


\bibitem[\protect\citeauthoryear{Qin, Chen, and Zhu}{Qin et~al\mbox{.}}{2014}]%
        {Qin2014Contextual}
\bibfield{author}{\bibinfo{person}{Lijing Qin}, \bibinfo{person}{Shouyuan
  Chen}, {and} \bibinfo{person}{Xiaoyan Zhu}.} \bibinfo{year}{2014}\natexlab{}.
\newblock \showarticletitle{Contextual Combinatorial Bandit and its Application
  on Diversified Online Recommendation}. In \bibinfo{booktitle}{\emph{SDM}}.
  \bibinfo{pages}{461--469}.
\newblock


\bibitem[\protect\citeauthoryear{Qin and Zhu}{Qin and Zhu}{2013}]%
        {Qin2013-promoting}
\bibfield{author}{\bibinfo{person}{Lijing Qin} {and} \bibinfo{person}{Xiaoyan
  Zhu}.} \bibinfo{year}{2013}\natexlab{}.
\newblock \showarticletitle{Promoting Diversity in Recommendation by Entropy
  Regularizer}. In \bibinfo{booktitle}{\emph{IJCAI}}.
  \bibinfo{pages}{2698--2704}.
\newblock


\bibitem[\protect\citeauthoryear{Radlinski, Kleinberg, and Joachims}{Radlinski
  et~al\mbox{.}}{2008}]%
        {radlinski08learning}
\bibfield{author}{\bibinfo{person}{Filip Radlinski}, \bibinfo{person}{Robert
  Kleinberg}, {and} \bibinfo{person}{Thorsten Joachims}.}
  \bibinfo{year}{2008}\natexlab{}.
\newblock \showarticletitle{Learning Diverse Rankings with Multi-Armed
  Bandits}. In \bibinfo{booktitle}{\emph{ICML}}. \bibinfo{pages}{784--791}.
\newblock


\bibitem[\protect\citeauthoryear{Slivkins, Radlinski, and Gollapudi}{Slivkins
  et~al\mbox{.}}{2010}]%
        {slivkins2010learning}
\bibfield{author}{\bibinfo{person}{Alex Slivkins}, \bibinfo{person}{Filip
  Radlinski}, {and} \bibinfo{person}{Sreenivas Gollapudi}.}
  \bibinfo{year}{2010}\natexlab{}.
\newblock \showarticletitle{Learning Optimally Diverse Rankings over Large
  Document Collections}. In \bibinfo{booktitle}{\emph{ICML}}.
  \bibinfo{pages}{983--990}.
\newblock


\bibitem[\protect\citeauthoryear{Thompson}{Thompson}{1933}]%
        {thompson-1933}
\bibfield{author}{\bibinfo{person}{William~R. Thompson}.}
  \bibinfo{year}{1933}\natexlab{}.
\newblock \showarticletitle{On the Likelihood that One Unknown Probability
  Exceeds Another in View of the Evidence of Two Samples}.
\newblock \bibinfo{journal}{\emph{Biometrika}}  \bibinfo{volume}{25}
  (\bibinfo{year}{1933}), \bibinfo{pages}{285--294}.
\newblock


\bibitem[\protect\citeauthoryear{Yue and Guestrin}{Yue and Guestrin}{2011}]%
        {yue-2011-linear}
\bibfield{author}{\bibinfo{person}{Yisong Yue} {and} \bibinfo{person}{Carlos
  Guestrin}.} \bibinfo{year}{2011}\natexlab{}.
\newblock \showarticletitle{Linear Submodular Bandits and their Application to
  Diversified Retrieval}. In \bibinfo{booktitle}{\emph{NIPS}}.
  \bibinfo{pages}{2483--2491}.
\newblock


\bibitem[\protect\citeauthoryear{Zoghi, Tunys, Ghavamzadeh, Kveton, Szepesvari,
  and Wen}{Zoghi et~al\mbox{.}}{2017}]%
        {batchrank}
\bibfield{author}{\bibinfo{person}{Masrour Zoghi},
  \bibinfo{person}{Tom{\'a}{\v{s}} Tunys}, \bibinfo{person}{Mohammad
  Ghavamzadeh}, \bibinfo{person}{Branislav Kveton}, \bibinfo{person}{Csaba
  Szepesvari}, {and} \bibinfo{person}{Zheng Wen}.}
  \bibinfo{year}{2017}\natexlab{}.
\newblock \showarticletitle{Online Learning to Rank in Stochastic Click
  Models}. In \bibinfo{booktitle}{\emph{ICML}}. \bibinfo{pages}{4199--4208}.
\newblock


\bibitem[\protect\citeauthoryear{Zoghi, Tunys, Li, Jose, Chen, Chin, and
  de~Rijke}{Zoghi et~al\mbox{.}}{2016}]%
        {zoghi-click-2016}
\bibfield{author}{\bibinfo{person}{Masrour Zoghi},
  \bibinfo{person}{Tom{\'a}{\v{s}} Tunys}, \bibinfo{person}{Lihong Li},
  \bibinfo{person}{Damien Jose}, \bibinfo{person}{Junyan Chen},
  \bibinfo{person}{Chun~Ming Chin}, {and} \bibinfo{person}{Maarten de Rijke}.}
  \bibinfo{year}{2016}\natexlab{}.
\newblock \showarticletitle{Click-based Hot Fixes for Underperforming Torso
  Queries}. In \bibinfo{booktitle}{\emph{SIGIR}}. \bibinfo{pages}{195--204}.
\newblock


\bibitem[\protect\citeauthoryear{Zong, Ni, Sung, Ke, Wen, and Kveton}{Zong
  et~al\mbox{.}}{2016}]%
        {zong16cascading}
\bibfield{author}{\bibinfo{person}{Shi Zong}, \bibinfo{person}{Hao Ni},
  \bibinfo{person}{Kenny Sung}, \bibinfo{person}{Nan~Rosemary Ke},
  \bibinfo{person}{Zheng Wen}, {and} \bibinfo{person}{Branislav Kveton}.}
  \bibinfo{year}{2016}\natexlab{}.
\newblock \showarticletitle{Cascading Bandits for Large-Scale Recommendation
  Problems}. In \bibinfo{booktitle}{\emph{UAI}}.
\newblock


\end{thebibliography}
